\documentclass{article}

\usepackage{arxiv}

\usepackage[utf8]{inputenc}
\usepackage{hyperref}
\usepackage{algorithm}
\usepackage{algpseudocode}
\usepackage{multirow}
\usepackage{url}
\usepackage{natbib}
\usepackage{physics}
\usepackage{xcolor}
\usepackage{amsmath, amsthm, amsfonts}
\usepackage{graphicx}
\usepackage{thmtools,thm-restate}
\usepackage{subcaption}

\DeclareMathOperator*{\supremum}{sup}
\DeclareMathOperator*{\argmin}{argmin}
\theoremstyle{plain}
\newtheorem{theorem}{Theorem}[section]

\theoremstyle{definition}
\newtheorem{definition}[theorem]{Definition}
\newtheorem{assumption}[theorem]{Assumption}
\theoremstyle{remark}
\newtheorem{remark}[theorem]{Remark}
\newtheorem{sublemma}{Lemma}[subsection]
\newtheorem{subcorollary}[sublemma]{Corollary}

\title{PINN-BO: A Black-box Optimization Algorithm using Physics-Informed Neural Networks}

\author{ Dat Phan-Trong \\
	Applied Artificial Intelligence Institute\\
	Deakin University\\
	Waurn Ponds, VIC 3216 \\
	\texttt{trongp@deakin.edu.au} \\
	\And
	Hung The Tran \\
	Applied Artificial Intelligence Institute\\
	Deakin University\\
	Waurn Ponds, VIC 3216 \\
	\texttt{hung.tranthe@deakin.edu.au} \\
    \And
	Alistair Shilton \\
	Applied Artificial Intelligence Institute\\
	Deakin University\\
	Waurn Ponds, VIC 3216 \\
	\texttt{alistair.shilton@deakin.edu.au} \\
    \\
    \And
	Sunil Gupta \\
	Applied Artificial Intelligence Institute\\
	Deakin University\\
	Waurn Ponds, VIC 3216 \\
	\texttt{sunil.gupta@deakin.edu.au} \\
}

\begin{document}
\allowdisplaybreaks
\raggedbottom
\maketitle

\begin{abstract}
	Black-box optimization is a powerful approach for discovering global optima in noisy and expensive black-box functions, a problem widely encountered in real-world scenarios. Recently, there has been a growing interest in leveraging domain knowledge to enhance the efficacy of machine learning methods. Partial Differential Equations (PDEs) often provide an effective means for elucidating the fundamental principles governing the black-box functions. In this paper, we propose PINN-BO, a black-box optimization algorithm employing Physics-Informed Neural Networks that integrates the knowledge from Partial Differential Equations (PDEs) to improve the sample efficiency of the optimization. We analyze the theoretical behavior of our algorithm in terms of regret bound using advances in NTK theory and prove that the use of the PDE alongside the black-box function evaluations, PINN-BO leads to a tighter regret bound. We perform several experiments on a variety of optimization tasks and show that our algorithm is more sample-efficient compared to existing methods.
\end{abstract}


\section{Introduction}
\label{PINN-BO:intro}

Black-box optimization has emerged as an effective technique in many real-world applications to find the global optimum of expensive, noisy black-box functions. Some notable applications include hyper-parameter optimization in machine learning algorithms \cite{snoek2012practical,bergstra2012random}, synthesis of short polymer fiber materials, alloy design, 3D bio-printing, and molecule design \cite{greenhill2020bayesian,shahriari2015taking}, optimizing design parameters in computational fluid dynamics \cite{morita2022applying}, and scientific research (e.g., multilayer nanoparticle,  photonic crystal topology) \cite{kim2022deep}. Bayesian Optimization is a popular example of black-box optimization method. Typically, Bayesian Optimization algorithms use a probabilistic regression model, such as a Gaussian Process (GP), trained on existing function observations. This model is then utilized to create an acquisition function that balances exploration and exploitation to recommend the next evaluation point for the black-box functions. Various options exist for acquisition functions, including improvement-based methods like Probability of Improvement \cite{kushner1964new}, Expected Improvement \cite{mockus1978application}, the Upper Confidence Bound \cite{srinivas2009gaussian}, Entropy Search \cite{hennig2012entropy} \cite{wang2017max}, Thompson Sampling \cite{chowdhury2017kernelized}, and Knowledge Gradient \cite{frazier2008knowledge}.

In the realm of objective functions encountered in scientific and engineering domains, many are governed by Partial Differential Equations (PDEs). These equations encapsulate the fundamental laws of physics that describe how systems evolve over time and space. For example, the heat equations describe the distribution of heat in a given space over time, the Navier-Stokes equations describe how the velocity, pressure, and density of a fluid change over time and space. They take into account factors such as the viscosity (resistance to flow) of the fluid and external forces acting upon it. Furthermore, physical laws being implied in PDE can also be found in structural analysis or electromagnetics, to name a few. Notably, recent efforts have been made to integrate PDE knowledge into objective function models. Raissi et al. introduce a new method using Gaussian Processes Regression (GPR) with a unique four-block covariance kernel in \cite{raissi2017machine}, enabling the utilization of observations from both the objective function and the Partial Differential Equations (PDEs). Another approach, as described in \cite{jidling2017linearly}, proposes a specialized covariance kernel to constrain Gaussian processes using differential equations. Unlike the method in \cite{raissi2017machine}, this approach enforces the constraint globally instead of relying on specific data points. This not only provides a stronger constraint but also eliminates the computational burden associated with the four-block covariance matrix. For more details, see \cite{swiler2020survey}. Recently, \cite{chen2021solving} proposed a numerical algorithm to approximate the solution of a given non-linear PDE as a Maximum a Posteriori (MAP) estimator of a Gaussian process conditioned on a finite set of data points from the PDE. Remarkably, their approach offers guaranteed convergence for a broad and inclusive class of PDEs. Despite the promising potential of Gaussian processes (GPs) in solving Partial Differential Equations (PDEs), GPs have a limitation as their computational scalability is a critical problem. The kernel matrix inversion when updating the posterior of a GP exhibits cubic complexity in the number of data points.

Recently, the Physics-Informed Neural Network (PINN) was introduced \cite{raissi2019physics, yang2021b}, which offers a viable approach for tackling general PDEs. Unlike GP, PINN leverages the expressive power of neural networks to approximate complex, nonlinear relationships within the data. This inherent flexibility enables PINN to handle a broader range of PDEs including nonlinear PDEs, making them well-suited for diverse scientific and engineering applications. Further, there is research providing a deeper insight into the theoretical aspect of incorporating the PDEs using PINN \cite{schiassi2021extreme, wangL2}, and analyzing the connection between GP and PINN models to learn the underlying functions satisfy PDE equations \cite{wang2022and}. 

As the PDEs hold promise as a valuable source of information, they could greatly improve the modeling of the black-box function and thus help towards sample-efficient optimization, reducing the number of function evaluations. It is worth noting that there has been relatively little exploration of how PDEs can be leveraged in the context of black-box optimization settings, where the objective function is treated as an unknown and potentially noisy function. 
In this paper, we consider a global optimization problem setting where the objective function $f \colon \mathcal{D} \rightarrow \mathbb{R}$ is associated with a PDE:   
\begin{equation*}
        \underset{\mathbf{x} \in \mathcal{D}}{\min} f(\mathbf{x}) \text{ s.t. }  \mathcal{N}[f](\mathbf{x}) = g(\mathbf{x}),
\end{equation*} 
where $\mathcal{D} \subset \mathbb{R}^d$ is a $d$-dimensional bounded domain and $\mathcal{N}[f]$ denotes a differential operator of the function $f$ with respect to the input $\mathbf{x}$. The function $f$ is an expensive, black-box function, and its evaluations are obtainable only through noisy measurements in the form of $y = f(\mathbf{x}) + \epsilon$, where $\epsilon$ represents sub-Gaussian noise, as elaborated later in Section \ref{section:theoretical_analysis}. Additionally, the function $g(\mathbf{x})$ is a cheap-to-evaluate function, which may also involve noise, with respect to the PDE-constraint. Furthermore, it is assumed that the boundary conditions of the PDEs are either unknown or inaccessible. These assumptions widely hold in many problem settings. As an example, \citet{cai2020heat} examines a two-dimensional heat transfer problem with forced heat convection around a circular cylinder. The heat measurement entails high costs due to the material, size, and shape of the system. The problem has known incompressible Navier-Stokes and heat transfer equations. However, the thermal boundary conditions are difficult to ascertain precisely because of the complex and large instruments.

To solve the aforementioned problem, we introduce a black-box optimization algorithm that employs a physics-informed neural network to model the unknown function.  Our approach adopts a straightforward greedy strategy to suggest the next function evaluation point. The neural network is trained by incorporating observations obtained from the black-box function and data derived from partial differential equations (PDEs) in the loss function, ensuring a comprehensive learning process.  Furthermore, we leverage recent advancements in neural network theory, specifically the Neural Tangent Kernel of Physics-Informed Neural Network (NTK-PINN), to analyze the theoretical behavior of our algorithm in the context of an infinite-width network setting. Our analysis includes demonstrating the convergence of our algorithm in terms of its regret bound, and showcasing its sample efficiency when compared to existing methods. Our contributions are summarized as:

\begin{itemize}
    \item We introduce a novel black-box optimization problem with physics information, described by Partial Differential Equations (PDEs), which is used to govern the objective function.
    \item We propose PINN-BO, a black-box optimization algorithm employing Physics-Informed Neural Networks with PDEs induced by natural laws to perform efficient optimization, bringing several benefits: improved sample-efficiency of optimization, scalable computation that only grows linearly with the number of function evaluations, and the ability to incorporate a broad class of PDEs (e.g. linear and non-linear).
    \item We provide a theoretical analysis of our proposed PINN-BO algorithm to illustrate that incorporating PDEs can lead to $\mathcal{O}\left(\sqrt{T\gamma_T}\sqrt{\gamma_T - I (f; \mathbf{Y}_T; \mathbf{U}_r) } \right)$ regret, where $T$ is the number of black-box function evaluations and $I (f; \mathbf{Y}_T; \mathbf{U}_r)$ is interaction information between the black-box function  $f$, its observations $\mathbf{Y}_T$ and the PDE data $\mathbf{U}_r$ (see Section \ref{section:theoretical_analysis}).
    \item We perform experiments with a variety of tasks showing that our algorithm outperforms current state-of-the-art black-box optimization methods.
\end{itemize} 
\section{Related Works}
\label{section:related_works}
In recent times, the application of Machine Learning to scientific fields, known as ``Machine Learning for Science'', has gained significant attention. This approach leverages the advancements in Machine Learning to tackle complex problems encountered in fundamental sciences, such as physics and chemistry. In the realm of natural sciences, particularly in physics, Partial Differential Equations (PDEs) frequently serve as an effective framework for describing the underlying principles governing various phenomena. Physics-Informed Neural Networks (PINNs) are a notable development in this domain. Unlike traditional approaches that solely rely on data to infer solutions, PINNs incorporate the underlying physics of the problem encoded by the PDE itself. The concept of building physics-informed learning machines, which systematically integrate prior knowledge about the solution, has its roots in earlier work by \cite{owhadi2015bayesian}, showcasing the promise of leveraging such prior information. In the research conducted by \cite{raissi2017inferring, raissi2017machine}, Gaussian process regression was employed to construct representations of linear operator functionals. This approach accurately inferred solutions and provided uncertainty estimates for a range of physical problems. Subsequently, this work was expanded upon in \cite{raissi2018numerical,raissi2018hidden}. In 2019, PINNs were introduced as a novel class of data-driven solvers \cite{raissi2019physics}. This groundbreaking work introduced and demonstrated the PINN approach for solving nonlinear PDEs, such as the Schrödinger, Burgers, and Allen–Cahn equations. PINNs were designed to address both forward problems, where solutions of governing mathematical models are estimated, and inverse problems, wherein model parameters are learned from observable data.

There is a line of research that utilizes derivatives information of objective function $f$ to improve the accuracy of the surrogate model and guide the optimization. \cite{lizotte2008practical} conducted empirical research on variants of EI algorithms incorporating derivative information. They demonstrated that Bayesian optimization utilizing the expected improvement (EI) acquisition function and full gradient information at each sample can outperform BFGS. In \cite{osborne2009gaussian}, derivative observations are employed as an alternative to function observations to improve the conditioning of the covariance matrix. Specifically, when sampling near previously observed points, only derivative information is used to update the covariance matrix. \cite{wu2017bayesian} introduced a derivative-based knowledge gradient (d-KG) algorithm and established its asymptotic consistency as well as its one-step Bayes optimality. \cite{penubothula2021novel} introduced a  first-order Bayesian Optimization approach that capitalizes on a fundamental insight: the gradient reaches zero at maxima, and demonstrated the enhanced performance and computational efficiency through numerical experiments compared to existing Bayesian Optimization methods on standard test functions. \cite{shekhar2021significance} proposed a first-order Bayesian Optimization (FOO) algorithm with two stages. The first stage identifies an active region that contains a neighborhood of presumed optimum, while the second stage involves taking local gradient steps based on estimates of the true gradients achieved through querying. The algorithm is proved to achieve a regret bound of $\mathcal{O}(d \log^2 T)$, where $d$ is the dimensionality of the input and $T$ is the number of observations. However, to the best of our knowledge,  no attempts have been made to incorporate the information given by partial derivative equations (PDEs) to enhance the performance of optimization problems under a black-box setting. 
\section{Proposed Method}
\label{section:method}
\label{PINN-BO:method}
In this section, we present our proposed Physics-Informed Neural Network based Black-box Optimization (PINN-BO). PINN-BO algorithm combines optimization and machine learning techniques to efficiently optimize an unknown black-box function over a given input space while leveraging physics-informed constraints described by a PDE. Following the fundamental principles of Bayesian optimization, our algorithm consists of two primary steps: (1) Constructing a model of the black-box objective function, and (2) Employing this model to select the next function evaluation point during each iteration. In the first step, our approach diverges from traditional Bayesian Optimization algorithms that typically utilize Gaussian processes to model the objective function. Instead, we employ a fully connected neural network denoted as $h(\mathbf{x}; \boldsymbol{\theta})$ to learn the function $f$ as follows:
\[
h(\mathbf{x};\boldsymbol{\theta}) = \frac{1}{\sqrt{m}} \mathbf{W}_L \phi(\mathbf{W}_{L-1}\phi(\cdots\phi(\mathbf{W}_1 \mathbf{x})),
\]
where $\phi\colon \mathbb{R} \rightarrow \mathbb{R}$ is a coordinate-wise smooth activation function (e.g., ReLU, Tanh), $\mathbf{W}_1 \in \mathbb{R}^{m \times d}, \mathbf{W}_i \in \mathbb{R}^{m \times m}, 2\leq i \leq L-1, \mathbf{W}_L \in \mathbb{R}^{1 \times m}$, and $\boldsymbol{\theta} \in \mathbb{R}^p$ is the collection of parameters of the neural network, $p=md+m^2(L-2)+m$ and $d$ is the dimension of inputs, i.e., $\mathbf{x} \in \mathcal{D} \subset \mathbb{R}^d$. We initialize all the weights to be independent and identically distributed as standard normal distribution $\mathcal{N}(0,1)$ random variables. To leverage the information embedded within the partial differential equation (PDE) governing the objective function $f$, our algorithm generates a set of $N_r$ PDE data points denoted as $\mathcal{R} =  \{\mathbf{z}_j, u_j\}_{j=1}^{N_r}$. Here, $u_j$ represents the noisy evaluations of the function $g$ at the corresponding point $\mathbf{z}_j$, where $u_j = g(\mathbf{z}_j) + \eta_j$. Besides, we denote $\mathcal{D}_t = \{\mathbf{x}_i, y_i\}_{i=1}^t$ as the set of noisy observations of the unknown function $f$ after $t$ optimization iterations, where $y_t = f(\mathbf{x}_t) + \epsilon_t$. We further define some other notations:
\begin{equation*}
        \phi(\cdot) = \nabla_{\boldsymbol{\theta}} h(\cdot; \boldsymbol{\theta}_0) ; \; \; \; \;
        \omega(\cdot)  = \nabla_{\boldsymbol{\theta}} \mathcal{N}[h] (\cdot; \boldsymbol{\theta}_0)\\
\end{equation*}
where $\phi(\cdot)$ is the gradient of $h(\cdot; \boldsymbol{\theta}_0)$ with respect to the parameter $\boldsymbol{\theta}$, evaluated at initialization $\boldsymbol{\theta}_0$. Similarly, $\omega(\cdot)$ represents the gradient of $\mathcal{N}[h](\cdot; \boldsymbol{\theta}_0)$ with respect to model parameters $\boldsymbol{\theta}$ and evaluated at initialization $\boldsymbol{\theta}_0$, where $\mathcal{N}[h](\cdot; \boldsymbol{\theta}_0)$ is the result of applying differential operator $\mathcal{N}$ (with respect to the input) to $h(\cdot; \boldsymbol{\theta}_0)$. Both $\mathcal{D}_t$ and $\mathcal{R}$ play an important role in the subsequent stages of the algorithm, specifically in the minimization of the loss function associated with learning the network $h(\mathbf{x}, \boldsymbol{\theta}_t)$ at optimization iteration $t$:
\begin{equation}
    \label{eqn:pinn_loss}
    \resizebox{0.48\textwidth}{!}{
$\mathcal{L}(t) = \sum^{t-1}_{i=1} [y_i - \nu_t h(\mathbf{x}_i; \boldsymbol{\theta}_{t-1})]^2 + \sum^{N_r}_{j=1}[u_j - \nu_t \mathcal{N}[h](\mathbf{z}_j; \boldsymbol{\theta}_{t-1})]^2$,
}
\end{equation}
where $\nu_t$ is a scale parameter that controls the exploration-exploitation trade-off. 

For the second step, we employ a greedy strategy to pick the next sample point $\mathbf{x}_t$. At each iteration $t$, the algorithm updates the neural network by optimizing the loss function described in Eqn \ref{eqn:pinn_loss} by gradient descent, with scaled function value predictions $\nu_t h(\cdot; \boldsymbol{\theta}_{t-1})$ and scaled predictions with respect to the governed PDE $\nu_t \mathcal{N}[h](\cdot; \boldsymbol{\theta}_{t-1})$. In the proof of Section \ref{section:theoretical_analysis}, we show this action is equivalent to placing the GP prior over function values $f_{1:t}$ and PDE values $g_{1:N_r}$ (Corollary \ref{corollary:PINN_GP_func}). Then the posterior distribution of the function prediction $\widetilde {f}_t(\mathbf{x}) = h(\mathbf{x}, \boldsymbol{\theta}_{t-1})$ at a new data point $\mathbf{x}$ can be viewed as being sampled from a GP with specific posterior mean and variance function (Lemma \ref{lemma:PINN_mean_cov}). This allows us to directly use the network prediction as an acquisition function following the principle of Thompson Sampling.  
Then, the next evaluation point $\mathbf{x}_t$ is selected by minimizing this acquisition function $\widetilde {f}_t(\mathbf{x}) = h(\mathbf{x}, \boldsymbol{\theta}_{t-1})$. Then, the black-box function is queried at point $\mathbf{x}_t$, resulting in a (noisy) observation $y_t$, which is subsequently used to update the dataset $\mathcal{D}_t$. The PDE observations set $\mathcal{R}$, in combination with the observations in $\mathcal{D}_t$, is integrated into the training process of the neural network by minimizing the squared loss, as described in Equation \ref{eqn:pinn_loss}. We provide a concise step-by-step summary of our approach in Algorithm \ref{alg:PINN-BO}.

To enhance the exploration step, it is important to bring the additional information to improve our model of objective function, especially in the regions where optima lies. In our case, this task of exploration is easier as we have access to PDE which provides knowledge about the objective function and reduces the amount of information that is needed to model the function. In section \ref{section:theoretical_analysis} (Theoretical Analysis), we derive a scaling factor $\nu_t = \widetilde{R} \sqrt{2\gamma_t - 2 I(f; \mathbf{Y}_t; \mathbf{U}_r)  + \log(\frac{1}{\delta})}$,  which directly reflects this intuition. It clearly reduces the maximum information gain  (which can be thought of as the complexity of the function modeling) by the interaction information $I(f; \mathbf{Y}_t; \mathbf{U}_r)$, which is a generalization of the mutual information for three variables: unknown function $f$,  its observations $\mathbf{Y}_t$, and the PDE data $\mathbf{U}_r$. This information can be calculated as $I(f; \mathbf{Y}_t; \mathbf{U}_r) = \frac{1}{2}  \log (\frac{\det(\frac{\Phi_t^\top \Phi_t}{\lambda_1} + \mathbf{I})\det(\frac{\Omega_r^\top \Omega_r}{\lambda_2} + \mathbf{I})}{\det(\frac{\Phi_t^\top \Phi_t}{\lambda_1} + \frac{\Omega_r^\top \Omega_r}{\lambda_2} + \mathbf{I})}) $, where $\Phi_t  = [\phi(\mathbf{x}_1)^\top,\dots, \phi(\mathbf{x}_t)^\top ] ^\top$ and 
    $\Omega_r  = [\omega(\mathbf{z}_1)^\top, \dots, \omega(\mathbf{z}_{N_r})^\top ] ^\top$. The value of $\nu_t$ signifies how our algorithm continues the exploration in regions of search space where the function $f$ has no implicit knowledge through PDE observations. These are the regions indicated by $\gamma_t - I(f; \mathbf{Y}_t; \mathbf{U}_r)$, which is the amount of information about the unknown function $f$ remains after our algorithm interacts with PDE data $\mathbf{U}_r$.

\begin{algorithm}[!ht]
\caption{Physics-informed Neural Network based Black-box optimization (PINN-BO)}
\label{alg:PINN-BO}
\textbf{Input}: The input space $\mathcal D$, the optimization budget $T$, PDE training set size $N_r$.
\begin{algorithmic}[1]
\State Initialize $\mathcal{D}_0 = \emptyset$ and $\boldsymbol{\theta_0} \sim \mathcal{N}(0,\mathbf{I})$ 
\State Generate set $\mathcal{R} = \{\mathbf{z}_j, u_j\}_{j=1}^{N_r}$ from the PDE.

\For{$t = 1$ to $T$}
\State Set $\nu_t = \widetilde{R} \sqrt{2\gamma_t - 2 I(f; \mathbf{Y}_t; \mathbf{U}_r)  + \log(\frac{1}{\delta})}$, where $\widetilde{R} = \sqrt{\left(\frac{R_1}{\lambda_1}\right)^2 + \left(\frac{R_2}{\lambda_2}\right)^2}$. 
\State $\widetilde{f_t}(\mathbf{x}) = h(\mathbf{x}; \boldsymbol{\theta}_{t-1})$ 
\State Choose $\mathbf{x}_t = \argmin_{\mathbf{x} \in \mathcal{D}} \widetilde{f_t}(\mathbf{x}) $ and receive observation $y_t = f(\mathbf{x}_t) + \epsilon_t$
\State Update $\mathcal{D}_t = \mathcal{D}_{t-1} \cup \{\mathbf{x}_t, y_t\}$
\State Update $ \boldsymbol{\theta}_t = \argmin_{\boldsymbol{\theta}} \mathcal{L}(\boldsymbol{\theta})$ using Eqn. \ref{eqn:pinn_loss} by gradient descent with $\nu = \nu_t$.
\EndFor
\end{algorithmic}
\end{algorithm}
\section{Theoretical Analysis}
\label{section:theoretical_analysis}
In this section, we provide a regret bound for the proposed PINN-BO algorithm. To quantify the algorithm's regret, we employ the cumulative regret, defined as $R_T = \sum_{t=1}^T r_t$ after $T$ iterations. Here, $\mathbf{x}^* = \argmin_{\mathbf{x} \in \mathcal{D}} f(\mathbf{x})$ represents the optimal point of the unknown function $f$, and $r_t = f(\mathbf{x^*}) - f(\mathbf{x}_t)$ denotes the instantaneous regret incurred at time $t$. Our regret analysis is built upon the recent NTK-based theoretical work of \cite{wang2022and} and proof techniques of GP-TS \cite{chowdhury2017kernelized}. Before proceeding with the theoretical analysis, we now introduce a set of definitions and assumptions. They clarify our proof and set up the basis and conditions for our analysis. \textbf{A detailed proof can be found in Section \ref{section:proof} of the Appendix.}   
\begin{definition}
    We define matrix $\mathbf{K}_\mathrm{NTK-PINN}$ as  the \textit{neural tangent kernel of a Physics-Informed Neural Network (NTK of PINNs)}: 
\begin{equation}
\renewcommand\arraystretch{1.2}
    \label{definition:PINN-NTKs}
\mathbf{K}_\mathrm{NTK-PINN} = 
\begin{bmatrix}
    \mathbf{K}_{uu} & \mathbf{K}_{ur} \\
    \mathbf{K}_{ru} & \mathbf{K}_{rr}
\end{bmatrix},
\end{equation}
where $(\mathbf{K}_{uu})_{ij} = \langle \phi(\mathbf{x}_i), \phi(\mathbf{x}_j) \rangle, 
 (\mathbf{K}_{ur})_{ij} = \langle \phi(\mathbf{x}_i), \omega(\mathbf{z}_j) \rangle, (\mathbf{K}_{rr})_{ij} = \langle \omega(\mathbf{z}_i), \omega(\mathbf{z}_j) \rangle$ and $\mathbf{K}_{ru} = \mathbf{K}_{ur}^\top$, defined using $\mathcal{D}_t$ and $\mathcal R$. 
\end{definition}
 

 \begin{assumption}
 \label{assumption:subgaussian}
 We assume the noises $\{\epsilon_i\}_{i=1}^T$ where $\epsilon_i = y_i - f(\mathbf{x}_i)$ and $\{\eta_j\}_{j=1}^{N_r}$ where $\eta_j = u_j - g(\mathbf{z}_j)$  are conditionally sub-Gaussian with parameter $R_1 > 0$ and $R_2 >0$, where $\{\epsilon_i\}_{i=1}^T$ and $\{\eta_j\}_{j=1}^{N_r}$ is assumed to capture the noises induced by querying the black-box, expensive function $f(\cdot)$   and cheap-to-evaluate PDE-related function $g(\cdot)$ defined in Section \ref{PINN-BO:intro}, respectively. 
 \begin{equation*}
     \begin{split}
         \forall i \ge 0, & \; \forall \lambda_1 \in \mathbb{R}, \;  \mathbb{E}[e^{\lambda_1\epsilon_i} \rvert \mathcal{F}_{t-1}] \le e^\frac{\lambda_1^2 R_1^2}{2} \\
         \forall j \ge 0, & \; \forall \lambda_2 \in \mathbb{R}, \;  \mathbb{E}[e^{\lambda_2\eta_j} \rvert \mathcal{F}_{N_r-1}^\prime] \le e^\frac{\lambda_2^2 R_2^2}{2}
     \end{split}
 \end{equation*}

 where $\mathcal{F}_{t-1}, \mathcal{F}_{N_r-1}^\prime$ are the $\sigma$-algebra generated by the random variables $\{\mathbf{x}_i, \epsilon_i\} 
^{t-1}_{i=1} \cup \{\mathbf{x}_t\}$ and $\{\mathbf{z}_j, \eta_j\} 
^{N_r-1}_{j=1} \cup \{\mathbf{z}_{N_r}\}$, respectively.
 \end{assumption}
 \begin{assumption}
    \label{assumption:rkhs}
     We assume $f$ to be an element of the Reproducing Kernel Hilbert Space (RKHS) associated with real-valued functions defined on the set $\mathcal{D}$. This specific RKHS corresponds to the Neural Tangent Kernel of a physics-informed neural network (NTK-PINN) and possesses a bounded norm denoted as $\norm{f}_{\mathcal{H}_{k_\textup{NTK-PINN}}} \leq B$.  Formally, this RKHS is denoted as $\mathcal{H}_{k_\textup{NTK-PINN}}(\mathcal D)$, and is uniquely characterized by its kernel function $k_\textup{NTK-PINN}(\cdot, \cdot)$. The RKHS induces an inner product $\langle \cdot, \cdot \rangle$ that obeys the reproducing property:
    $f(\mathbf{x}) = \langle f, k_\textup{NTK-PINN}(\cdot, \mathbf{x})\rangle$ for all $f \in  \mathcal{H}_{k_\textup{NTK-PINN}}(\mathcal{D})$. 
    The norm induced within this RKHS, $\norm{f}_{\mathcal{H}_{k_\textup{NTK-PINN}}} = \sqrt{\langle f,f\rangle_{\mathcal{H}_{k_\textup{NTK-PINN}}}}$, quantifies the smoothness of $f$ concerning the kernel function $k_\textup{NTK-PINN}$, and satisfies: $f \in \mathcal{H}_{k_\textup{NTK-PINN}}(\mathcal{D})$ if and only if $\norm{f}_{k_{\mathcal{H}_\textup{NTK-PINN}}} < \infty$. 
 \end{assumption}

Assumptions \ref{assumption:subgaussian} and \ref{assumption:rkhs} represent commonly employed and well-established assumptions in GP-based Bandits and Bayesian Optimization \cite{chowdhury2017kernelized, vakili2021optimal}. We are now prepared to establish an upper bound on the regret incurred by our proposed PINN-BO algorithm.  

We begin by presenting key lemmas for establishing the regret bound in Theorem \ref{theorem:regret_bound} of the proposed algorithms. The following lemma demonstrates that, given the assumption of an infinitely wide network, the output of trained physics-informed neural network after running $t$ optimization iterations in Algorithm \ref{alg:PINN-BO}, can be regarded as sampling from a GP with specific mean and covariance functions.
\begin{restatable}{lemma}{PinnMeanCov} 
\label{lemma:PINN_mean_cov}
Conditioned on $\mathcal{D}_t = \{\mathbf{x}_i, y_i\}_{i=1}^t, \mathcal{R} = \{\mathbf{z}_j, u_j\}_{j=1}^{N_r}$, the acquisition function $\widetilde{f}_t(\mathbf{x}) = h(\mathbf{x}; \boldsymbol{\theta}_{t-1})$ can be viewed as a random draw from a $\mathrm{GP}\left(\mu_{t}^f (\mathbf{x}), \nu_t^2 \left(\sigma^f_t\right)^2(\mathbf{x})\right)$ with the following mean and covariance functions:
\begin{equation*}
\begin{split}
     \mu_{t}^f (\mathbf{x}) &= \phi(\mathbf{x})^\top \boldsymbol{\xi}_t^\top \mathbf{\widehat{K}}_\mathrm{PINN}^{-1} 
     \renewcommand\arraystretch{1.2}
     \begin{bmatrix}
         \mathbf{Y}_t \\
         \mathbf{U}_r
     \end{bmatrix} \\
    \left(\sigma^f_t\right)^2(\mathbf{x}) &=  \langle \phi(\mathbf{x}),  \phi(\mathbf{x}) \rangle - \phi(\mathbf{x})^\top \boldsymbol{\xi}_t^\top \mathbf{\widehat{K}}_\mathrm{PINN}^{-1} \boldsymbol{\xi}_t \phi(\mathbf{x}), \\
\end{split}
\end{equation*}
where 
\begin{equation*}
    \renewcommand\arraystretch{1.2}
    \begin{split}
\mathbf{\widehat{K}}_\mathrm{PINN}^{-1} &=  
    \begin{bmatrix}
    \mathbf{K}_{uu} + \lambda_1 \mathbf{I} & \mathbf{K}_{ur} \\
    \mathbf{K}_{ru} & \mathbf{K}_{rr} + \lambda_2 \mathbf{I}
    \end{bmatrix} ^{-1}  = \begin{bmatrix}
    \widetilde{\mathbf{A}} & \widetilde{\mathbf{B}} \\
    \widetilde{\mathbf{C}} & \widetilde{\mathbf{D}}
    \end{bmatrix}\\
    \Phi_t & = [\phi(\mathbf{x}_1)^\top,\dots, \phi(\mathbf{x}_t)^\top ] ^\top \\
    \Omega_r & = [\omega(\mathbf{z}_1)^\top, \dots, \omega(\mathbf{z}_{N_r})^\top ] ^\top, \; \boldsymbol{\xi}_t = \begin{bmatrix}
    \Phi_t^\top & \Omega_r^\top  
    \end{bmatrix}^\top \\
    \mathbf{K}_{uu} &= \Phi_t \Phi_t^\top, \;  \mathbf{K}_{ur} = \Phi_t \Omega_r^\top,\mathbf{K}_{ru} = \mathbf{K}_{ur}^\top, \mathbf{K}_{rr} = \Omega_r \Omega_r^\top \\
    \mathbf{Y}_t &= [y_1, y_2, \dots, y_t]^\top, 
    \mathbf{U}_r = [u_1, u_2, \dots, u_{N_r}]^\top \\
    \end{split}
\end{equation*}
\end{restatable}

Generic BO methods (without the extra information from PDE) utilized the \textit{maximum information gain} over search space $\mathcal{D}$ at time $t$:
    $\gamma_t := \max_{\mathcal{A} \subset \mathcal{D}: \lvert \mathcal{A} \rvert=t} I(\mathbf{Y}_\mathcal{A}, f_\mathcal{A})$, 
    where $I(\mathbf{Y}_\mathcal{A}, f_\mathcal{A})$ denotes the mutual information between $f_\mathcal{A} = [f(\mathbf{x})]_{\mathbf{x}\in \mathcal{A}}$ and noisy observations $\mathbf{Y}_\mathcal{A}$, which quantifies the reduction in uncertainty about the objective function $f$ after observing $y_A$. The maximum information gain is the fundamental component when analyzing regret bound for their algorithm \cite{srinivas2009gaussian,vakili2021optimal}. However, in our work, the PDE evaluations $\{u_j\}_{j=1}^{N_r}$ of the function $g$ is considered as the second source of information that contributes to reduce the uncertainty of $f$. Therefore, we introduce the \textit{interaction information} as the generalization of \textit{mutual information} for three random variables:
\begin{definition}
\label{def:interaction_information}
The \textbf{interaction information} between $f$,  its observations $\mathbf{Y}_\mathcal{A} (\text{where } \mathcal{A} \subset \mathcal{D})$, and the PDE data $\mathbf{U}_r$ can be defined as:
\[
I (f; \mathbf{Y}_\mathcal{A}; \mathbf{U}_r) = I (f; \mathbf{Y}_\mathcal{A}) - I (f; \mathbf{Y}_\mathcal{A} \rvert \mathbf{U}_r),
\]
where $I(f; \mathbf{Y}_\mathcal{A})$ quantifies the reduction in uncertainty in $f$ due to observing $\mathbf{Y}_\mathcal{A}$, while $I(f; \mathbf{Y}_\mathcal{A} \rvert \mathbf{U}_r)$ represents the additional information contributed by $\mathbf{U}_r$ to enhance the mutual information between $f$ and $\mathbf{Y}_\mathcal{A}$. 
\end{definition}
The next lemma provides the closed-form expression of the interaction information. 
\begin{restatable}{lemma}{InteractionInformation} 
    \label{lemma:interaction_information_formula}
    The interaction information between $f$ and observation $\mathbf{Y}_t$ and PDE data $\mathbf{U}_r$, for the points chosen from Algorithm \ref{alg:PINN-BO} can be calculated as:
    \begin{equation*}
        \resizebox{0.48\textwidth}{!}{
        $I (f; \mathbf{Y}_t; \mathbf{U}_r) = \frac{1}{2}  \log (\frac{\det(\frac{\Phi_t^\top \Phi_t}{\lambda_1} + \mathbf{I})\det(\frac{\Omega_r^\top \Omega_r}{\lambda_2} + \mathbf{I})}{\det(\frac{\Phi_t^\top \Phi_t}{\lambda_1} + \frac{\Omega_r^\top \Omega_r}{\lambda_2} + \mathbf{I})})$
        }
    \end{equation*}
\end{restatable}

\begin{remark}
    \label{remark:non_negative_interaction_information}
    Following Remark 3.3 in \cite{wang2022and}, both matrices $\frac{\Phi_t^\top \Phi_t}{\lambda_1}$ and $\frac{\Omega_r^\top \Omega_r}{\lambda_2}$ are positive semi-definite. It can be clearly seen that the interaction information given in Lemma \ref{lemma:interaction_information_formula} is non-negative:
    \begin{equation*}
        \resizebox{0.48\textwidth}{!}{
        $\begin{aligned}
            I (f; \mathbf{Y}_t; \mathbf{U}_r) &= \frac{1}{2}  \log (\frac{\det(\frac{\Phi_t^\top \Phi_t}{\lambda_1} + \mathbf{I})\det(\frac{\Omega_r^\top \Omega_r}{\lambda_2} + \mathbf{I})}{\det(\frac{\Phi_t^\top \Phi_t}{\lambda_1} + \frac{\Omega_r^\top \Omega_r}{\lambda_2} + \mathbf{I})}) \\
            &= \frac{1}{2}  \log (\frac{\det(\frac{\Phi_t^\top \Phi_t}{\lambda_1} + \frac{\Omega_r^\top \Omega_r}{\lambda_2}  + \mathbf{I} + \frac{\Phi_t^\top \Phi_t \Omega_r^\top \Omega_r}{\lambda_1 \lambda_2})}{\det(\frac{\Phi_t^\top \Phi_t}{\lambda_1} + \frac{\Omega_r^\top \Omega_r}{\lambda_2} + \mathbf{I})}) \ge 0
        \end{aligned}$
        }
    \end{equation*}
    The inequality uses the identity $\det(\mathbf{A}+ \mathbf{B}) \ge \det(\mathbf{A})$, where $\mathbf{A},\mathbf{B}$ are two positive semi-definite matrices. 
\end{remark}
Our next result shows how the prediction of the neural network model is concentrated around the unknown reward function $f$, which is the key to a tighter regret bound.

\begin{restatable}{lemma}{ConfidenceBound}
\label{lemma:confidence_bound}
     Assume that $\norm{\omega(\cdot)}_2 \le L$, where $\omega(\cdot)  = \nabla_{\boldsymbol{\theta}} \mathcal{N}[h] (\cdot; \boldsymbol{\theta}_0)$ and $\rho_{min}(\mathbf{K}_{uu})$ the smallest eigenvalue of kernel matrix $\mathbf{K}_{uu}$ defined in lemma 1. Set $N_r = c_r\left(1+ \frac{\rho_{min}(\mathbf{K}_{uu})}{\lambda_1}\right)/L^2$ for a positive constant $c_r$. Under the same hypotheses as stated in Assumption \ref{assumption:subgaussian} and Assumption \ref{assumption:rkhs}, and denote $\widetilde{R} = \sqrt{\left(\frac{R_1}{\lambda_1}\right)^2 + \left(\frac{R_2}{\lambda_2}\right)^2}$. Let $\delta \in (0,1)$. Then,  with  probability at least $1 - \delta$, the following confidence bound holds for all $\mathbf{x} \in \mathcal{D}$ and $t \ge 1$:
    \begin{equation*}
    \resizebox{0.48\textwidth}{!}{
        $\begin{aligned}
            & \lvert f(\mathbf{x}) - \mu_t^f(\mathbf{x}) \rvert \\
            & \le \sigma_t^f(\mathbf{x}) \left(B +   \widetilde{R}\sqrt{2 I (f; \mathbf{Y}_t) - 2I (f; \mathbf{Y}_t; \mathbf{U}_r) + \mathcal{O}(1) + \log(1/\delta)}  \right) \\
            & \le \sigma_t^f(\mathbf{x}) \left(B +   \widetilde{R}\sqrt{2 \gamma_t - 2I (f; \mathbf{Y}_t; \mathbf{U}_r) + \mathcal{O}(1) + \log(1/\delta)}  \right)
        \end{aligned}$
        }
    \end{equation*}
\end{restatable}
\paragraph{Proof sketch for Lemma \ref{lemma:confidence_bound}}
We split the problem into two terms: The prediction error of an element $f$ in the RKHS as assumed in Assumption \ref{assumption:rkhs} with noise-free observations and the noise effect. Our proof differs from most GP-based Bayesian Optimization methods, which use single-block kernel matrices. In contrast, our predictive mean and covariance function involve the inversion of a block matrix  $\mathbf{\widehat{K}}_\mathrm{PINN}^{-1}$, as stated in Lemma \ref{lemma:PINN_mean_cov}. We employ the block matrix inversion formula (see Appendix A, \cite{rasmussen2006gaussian}) to express $\mathbf{\widehat{K}}_\mathrm{PINN}^{-1}$ as four distinct matrices. Subsequently, using equivalent transformations, intermediate matrix identities, and utilizing the expression \textit{Interaction information} provided in Lemma \ref{lemma:interaction_information_formula}, we derive the final bound. 

\begin{remark}
    The upper bound of the confidence interval presented in Lemma \ref{lemma:confidence_bound} shares a similar form with the existing confidence interval of GP-TS as outlined in \cite{chowdhury2017kernelized}. It is worth emphasizing, however, that our bound offers valuable insights into the significance of integrating partial differential equations (PDEs) to attain a tighter confidence bound. This insight can be summarized as follows: The expression $I(f; \mathbf{Y}_t) - I(f; \mathbf{Y}_t; \mathbf{U}_r)$ equals to $I(f; \mathbf{Y}_t|\mathbf{U}_r)$, which represents the expected mutual information between the function $f$ and the observations $\mathbf{Y}_t$, given $\mathbf{U}_r$. Lemma \ref{lemma:interaction_information_formula} quantifies $I(f; \mathbf{Y}_t; \mathbf{U}_r)$ in terms of the kernel Gram matrices induced by black-box function and the PDE observations. As mentioned in Remark \ref{remark:non_negative_interaction_information}, the condition $I(f; \mathbf{Y}_t; \mathbf{U}_r) \ge 0$ implies that $I(f; \mathbf{Y}_t) \ge I(f; \mathbf{Y}_t|\mathbf{U}_r)$.  This inequality signifies that knowing the values of the partial differential equation (PDE) component $\mathbf{U}_r$ reduces the statistical information between the observations $\mathbf{Y}_t$ and the unknown function $f$. In other words, knowing the values of PDE component $\mathbf{U}_r$ can diminish the number of observations $\mathbf{Y}_t$ required to estimate the unknown function $f$. 
\end{remark} 

\begin{remark}
The value of $I(f; \mathbf{Y}_t; \mathbf{U}r)$ depends on the specific problem. For instance, if we assume that $f$ is a function in RKHS with a linear kernel and $\mathcal{N}[f]= \sum_{i=1}^n \frac{\partial f}{\partial \mathbf{x}_i}$, the lower bound for the interaction information is: $I(f; \mathbf{Y}_t; \mathbf{U}_r) = \Theta\left(\frac{d N_r}{d N_r+1}(1-1/T)\right) = \Theta(1)$, which is a constant. This is because the linear kernel differential feature map sends all PDE points to the same vector in the RKHS. This example aims to show how to bound the interaction information for a known PDE. 
\end{remark}

\begin{restatable}{theorem}{RegretBound}
\label{theorem:regret_bound}
Let $\mathbf{K}_{uu}, \mathbf{K}_{ur}, \mathbf{K}_{ru},$ and $\mathbf{K}_{rr}$ be four matrices as defined in Lemma \ref{lemma:PINN_mean_cov}. 
Let $\delta \in (0,1)$. Assume that $\norm{\omega(\cdot)}_2 \le L$ and  $\rho_{min}(\mathbf{K}_{uu})$ be the smallest eigenvalue of kernel matrix $\mathbf{K}_{uu}$. Set $N_r = c_r\left(1+ \frac{\rho_{min}(\mathbf{K}_{uu})}{\lambda_1}\right)/L^2$ for a constant $c_r > 0$. Additionally,  let $\widetilde{R} = \sqrt{\left(\frac{R_1}{\lambda_1}\right)^2 + \left(\frac{R_2}{\lambda_2}\right)^2}$ and $I_0 = \frac{1}{2}\log \frac{\det(\mathbf{K}_{rr} + \lambda_2\mathbf{I})}{\det(\mathbf{K}_{rr} + \lambda_2\mathbf{I} - \mathbf{K}_{ru} \mathbf{K}_{uu}^{-1} \mathbf{K}_{ur})}$. Then with probability at least $1-\delta$, the regret of PINN-BO running for a function $f$ lying in the $\mathcal{H}_{k_\textup{NTK-PINN}}$, $\norm{f}_{{H}_{k_\textup{NTK-PINN}}} \le B$ as stated in Assumption \ref{assumption:rkhs}, after $T$ iterations satisfies:
\begin{equation*}
    \resizebox{0.48\textwidth}{!}{
    $\begin{aligned}
        R_T  = \mathcal{O} \Bigg(\sqrt{T d\log BdT} &\bigg[B \sqrt{\gamma_T - I_0 + \log(2/\delta)} 
        \\
        & + \widetilde{R} \sqrt{\gamma_T} \sqrt{\gamma_T - I (f; \mathbf{Y}_T; \mathbf{U}_r) - I_0 + \log(2/\delta
        )} \bigg] \Bigg)
    \end{aligned}$
    }
\end{equation*}
\end{restatable}
\section{Experimental Results}
\label{section:experiments}
In this section, we demonstrate the effectiveness of our proposed PINN-BO algorithm through its application of synthetic benchmark optimization functions as well as real-world optimization problems. 
\begin{figure*}[ht!] %
  \centering
  \includegraphics[width=\textwidth]{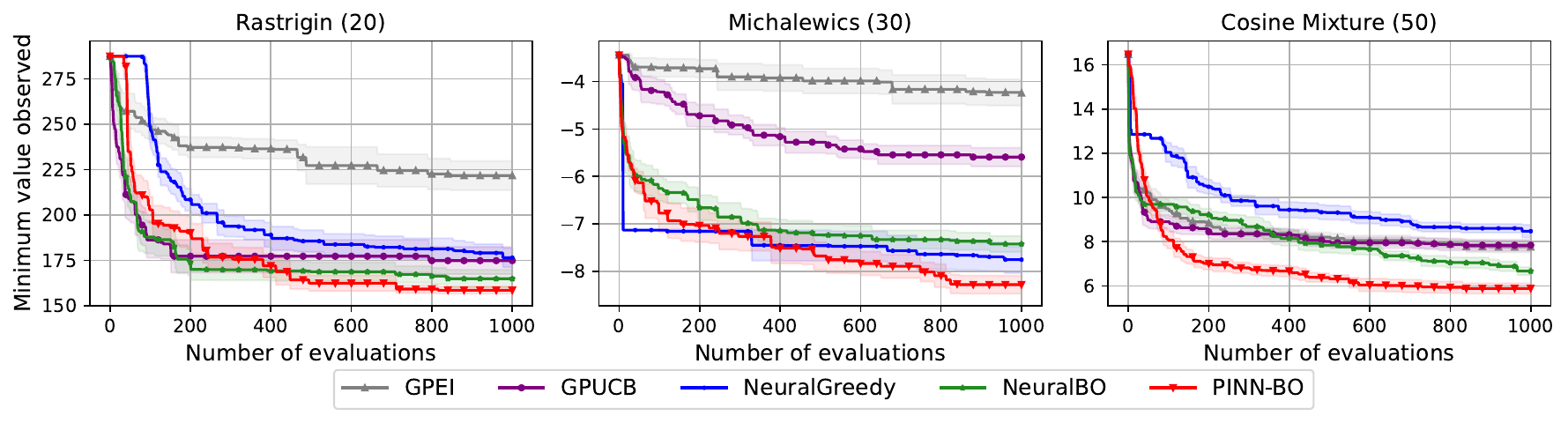} 
  \caption{The optimization results for synthetic functions comparing the proposed PINN-BO with the baselines. The standard errors are shown by color shading.}
  \label{fig:synthetic}
\end{figure*}
\subsection{Baselines}
\label{section:baselines}
For all experiments, we compared our algorithm with common classes of surrogate models used in black-box optimization, including Gaussian Processes (GPs) and Deep Neural Networks (DNNs). For GPs, we employ the most popular strategy GP-EI \cite{mockus1978application} and GP-UCB \cite{srinivas2009gaussian} with the Mat\'ern Kernel. Our implementations for GP-based Bayesian Optimization baselines utilize public library GPyTorch  \url{https://gpytorch.ai/} and BOTorch \url{https://botorch.org/}. We also include two recent DNNs-based works for black-box optimization: Neural Greedy \cite{pariagreedy} and NeuralBO \cite{PHANTRONG2023126776} described below:
\begin{itemize}
    \item NeuralGreedy \cite{pariagreedy} fits a neural network to the current set of observations, where the function values are randomly perturbed before learning the neural network. The learned neural network is then used as the acquisition function to determine the next query point. Since the NeuralGreedy code is not publicly available, we use our own implementation following the setting described in     \citet{pariagreedy} (see Appendix F.2 therein).
    \item NeuralBO \cite{PHANTRONG2023126776} utilizes the Thompson Sampling strategy for selecting the next evaluation point. In this approach, the mean function is estimated using the output of a fully connected deep neural network. To implement this baseline, we adhere to the configuration outlined in Section 7 in \citet{PHANTRONG2023126776}.
\end{itemize}
For our proposed PINN-BO algorithm, we employ a fully connected deep neural network (DNN) as the surrogate model. The network's weights are initialized with independent samples drawn from a normal distribution $\mathcal{N} (0, 1/m)$, where $m$ represents the width of the DNN. The model's hyper-parameters, which include depth, width, and learning rate, are selected as follows: For each function, we perform a grid search for tuning, where each hyper-parameter tuple is trained with 50 initial points. The width is explored within the set $\{100,200,500\}$, while the depth and the learning rate are searched across the values $\{2,3,4\}$ and $\{0.001, 0.005, 0.01, 0.02, 0.05, 0.1\}$, respectively.  Subsequently, we select the tuple of (depth, width, learning rate) associated with the lowest mean-square error during evaluation. To train the surrogate neural network models, we utilize the (stochastic) gradient descent optimizer along with an Exponential Learning Rate scheduler with a factor of $\gamma=0.95$. To accelerate the training process, we update the parameters $\boldsymbol{\theta}_t$ of surrogate models in Algorithm \ref{alg:PINN-BO} after every 10 optimization iterations with 100 epochs.  

\subsection{Synthetic Benchmark Functions}
 We conducted optimization experiments on five synthetic functions: DropWave (2), Styblinski-Tang (10), Rastrigin (20), Michalewics (30), and Cosine Mixture (50), where the numbers in parentheses indicate the input dimensions of each function. We selected them to ensure a diverse range of difficulty levels, as suggested by the difficulty rankings available at \url{https://infinity77.net/global_optimization/test_functions.html}. To enhance the optimization process, we incorporated the partial differential equations (PDEs) associated with these objective functions. The detailed expressions of these functions and their corresponding PDEs can be found in Section \ref{section:experiments_synthetic} of the  Appendix. Additionally, the noise in function evaluations follows a normal distribution with zero mean, and the variance is set to 1\% of the function range. Results for Rastrigin, Michalewics, and Cosine Mixture functions are presented in Figure \ref{fig:synthetic} in the main paper (the results for DropWave and Styblinski-Tang functions can be found in Section \ref{section:experiments_synthetic} of the Appendix).
All experiments reported here are averaged over 10 runs, each with random initialization. All methods begin with the same initial points. The results demonstrate that our PINN-BO is better than all other baseline methods, including GP-based BO algorithms (GP-EI, GP-UCB), and NN-based BO algorithms (NeuralBO, NeuralGreedy). 
\subsection{Real-world Applications}
In this section, we explore two real-world applications where the objective functions are constrained by specific partial differential equations (PDEs). We consider two tasks: (1) optimizing the Steady-State temperature distribution, satisfying the  Laplace equation, and (2) optimizing the displacement of a beam element, adhering to the non-uniform Euler-Bernoulli equation. We continue to compare our proposed method with the baselines mentioned in Section \ref{section:baselines}. Due to space constraints, we briefly introduce these problems in the following subsections and present figures illustrating the results of one case for each problem. Detailed experimental setup and other results are provided in Section \ref{section:experiments_real_world} of the  Appendix.
\subsubsection{Optimizing Steady-State Temperature}
\begin{figure}[ht]
  \centering  \includegraphics[width=0.48\textwidth]{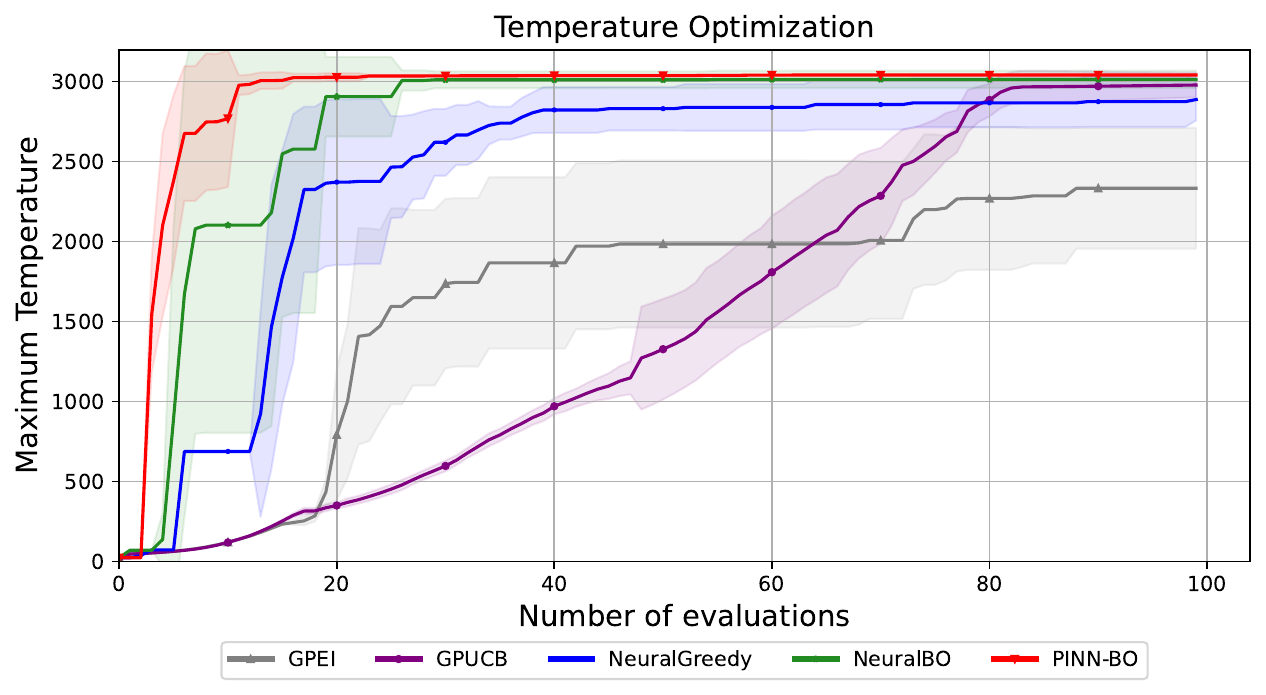}
  \caption{The optimization results for finding the maximum temperature comparing the proposed PINN-BO with the baselines. It can be seen that, using the PDE heat equation, PINN-BO found the maximum temperature faster than all baselines.}
  \label{fig:heat_main}
\end{figure}

The steady-state heat equation represents a special case of the heat equation when the temperature distribution no longer changes over time. It describes the equilibrium state of a system where the temperature is constant, and no heat is being added or removed. The steady-state heat equation is given by: $\nabla^2 T(x, y) = 0$, 
where $x, y$ are spatial variables that represent the positions within a two-dimensional space. In this paper, our task is to find the best position $(x,y)$ that maximizes temperature $T$ within the defined domain. Formally, this problem can be defined as: 
\begin{equation*}
        \underset{x,y \in \mathcal{U}}{\max}  \;T(x,y) \text{ s.t. }  \nabla^2 T(x, y) = 0,
\end{equation*}
We consider three different heat equations, where the solution of each problem is associated with one (unknown) boundary condition. The details and the results of optimizing these heat equation problems are provided in Section \ref{section:experiments_2d_laplace}. The optimization results of our PINN-BO, in comparison to other baseline methods, for the first case of the heat equation, are depicted in Figure \ref{fig:heat_main}.
\subsubsection{Optimizing Beam Displacement}
Euler-Bernoulli beam theory is a widely used model in engineering and physics to describe the behavior of slender beams under various loads. The governing differential equation for a non-uniform Euler-Bernoulli beam is given by: 
\begin{equation*}
    \frac{d^2}{dx^2} \left( EI(x) \frac{d^2 w(x)}{dx^2} \right) = q(x),
\end{equation*}
where
$EI(x)$ represents the flexural rigidity of the beam, which can vary with position $x$, and $w(x)$ represents the vertical displacement of the beam at position $x$, and $q(x)$ represents the distributed or concentrated load applied to the beam. The lower value deflection leads to a stiffer and more resilient structure, hence reducing the risk of structural failure and serviceability issues.  In this paper, we consider the task of finding the position $x$ that minimizes displacement $w(x)$. This problem can be defined as: 
\[ \underset{x \in \mathcal{S}}{\min}  \;
        w(x) \text{ s.t. }  \frac{d^2}{dx^2} \left( EI(x) \frac{d^2 w(x)}{dx^2} \right) = q(x)\] 
        
\begin{figure}[ht]
  \centering  \includegraphics[width=0.48\textwidth]{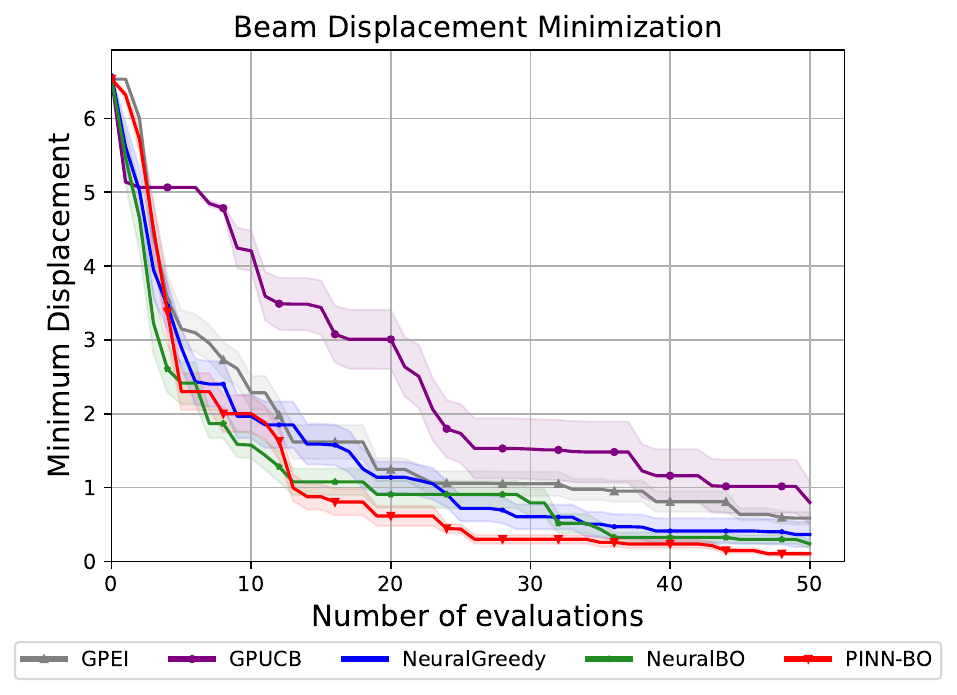}
  \caption{The minimum displacement on the non-uniform Euler beam under given loads $q(x)$,  flexural rigidity $EI(x)$, and boundary conditions. In comparison with other baselines, our proposed PINN-BO found the location with the smallest displacement, ensuring stability when placing the load over the beam.}
  \label{fig:beam_main}
\end{figure}
\section{Conclusion}
\label{section:conclusion}
We introduced a novel black-box optimization scenario incorporating Partial Differential Equations (PDEs) as additional information. Our solution, PINN-BO, is a new algorithm tailored to this challenge. We conducted a theoretical analysis, demonstrating its convergence with a tighter regret bound. Through experiments involving synthetic benchmark functions and real-world optimization tasks, we validated the efficacy of our approach.

\bibliographystyle{unsrtnat}
\bibliography{references}

\newpage
\appendix
\onecolumn
\section{Additional Experimental Results}
\subsection{Synthetic Benchmark Functions}
\label{section:experiments_synthetic}
We present the mathematical expressions of five synthetic objective functions and their accompanied PDEs used in Section \ref{section:experiments} of the main paper as follows: 
\paragraph{Drop-Wave:} 
\begin{align*}
    f(\mathbf{x}) &= - \frac{1 + \cos(12\sqrt{\mathbf{x}_1^2 + \mathbf{x}_2^2})}{0.5(\mathbf{x}_1^2 + \mathbf{x}_1^2) + 2} \text{ s.t. } \mathbf{x}_1 \frac{\partial f}{\partial \mathbf{x}_2} - \mathbf{x}_2 \frac{\partial f}{\partial \mathbf{x}_1} = 0
\end{align*}
\paragraph{Styblinski-Tang:}
\begin{align*}
        f(\mathbf{x}) &= \frac{1}{2} \sum_{i=1}^{d} (\mathbf{x}_i^4 - 16\mathbf{x}_i^2 + 5\mathbf{x}_i) \text{ s.t. }  \sum_{i=1}^d \frac{\partial f}{\partial \mathbf{x}_i} = \sum_{i=1}^{d}(2\mathbf{x}_i^3 -16\mathbf{x}_i +\frac{5}{2})
\end{align*}

\begin{figure*}[ht!] %
  \centering
  \includegraphics[width=\textwidth]{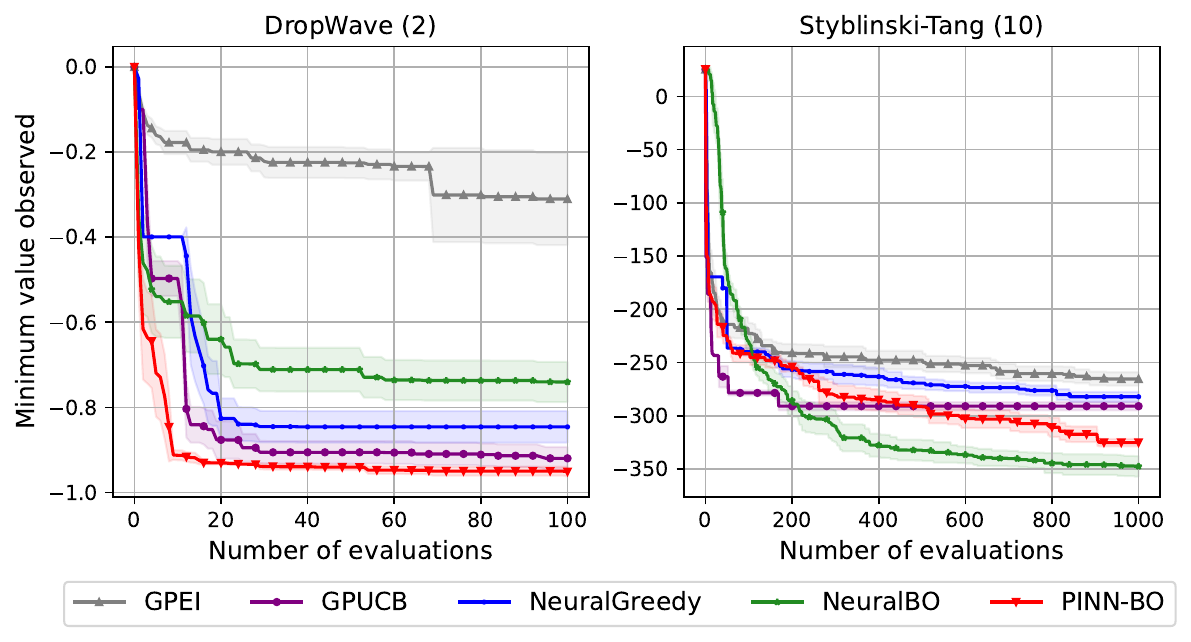} 
  \caption{The optimization results for synthetic functions comparing the proposed PINN-BO with the baselines. The standard errors are shown by color shading.}
  \label{fig:synthetic_supp}
\end{figure*}

\paragraph{Rastrigin:}
\begin{align*}
        f(\mathbf{x}) &= 10d + \sum_{i=1}^{d} \left[ \mathbf{x}_i^2 - 10 \cos(2\pi \mathbf{x}_i) \right] 
        \\
        \text{s.t. }  &\mathbf{x}^\top \nabla f(\mathbf{x})  - f(\mathbf{x}) = 10\sum_{i=1}^d \left[ (\cos(2\pi \mathbf{x}_i)) + \pi \mathbf{x}_i (\sin(2\pi \mathbf{x}_i)) - 1 \right]
\end{align*}
\paragraph{Michalewics:}
\begin{align*}
        f(\mathbf{x}) &= -\sum_{i=1}^{d} \sin(\mathbf{x}_i) \sin^{2m}\left(\frac{i\mathbf{x}_i^2}{\pi}\right) \text{ s.t. } \mathbf{h}^\top \nabla f(\mathbf{x}) - f(\mathbf{x}) = 0,\\ & \text{where } \mathbf{h} = [\mathbf{h}_1, \mathbf{h}_2, \dots, \mathbf{h}_d]^\top, \mathbf{h}_i = \left[\frac{\cos(\mathbf{x}_i)}{\sin(\mathbf{x}_i)} + \frac{2\mathbf{x}_i (2m-1)}{\tan(\frac{i\mathbf{x}_i^2}{\pi})}\right]^{-1}
\end{align*}
\paragraph{Cosine Mixture:}
\begin{align*}
    0.1 \sum_{i=1}^d \cos(5\pi\mathbf{x}_i) + \sum_{i=1}^d \mathbf{x}_i^2 \text{ s.t } \sum_{i=1}^d \left(\frac{\partial f}{\partial \mathbf{x}_i} - 2\mathbf{x}_i + 0.5\pi\sin(5\pi\mathbf{x}_i) \right)^2 = 0
\end{align*}
We show the optimization results of two remaining functions, including DropWave and Styblinski-Tang, optimized by our PINN-BO method and the other baselines, in Figure \ref{fig:synthetic_supp}.   
\subsection{Real-world Applications}
\label{section:experiments_real_world}
\subsubsection{Optimizing Steady-State Temperature}
\label{section:experiments_2d_laplace}
\begin{figure}[!ht]
    \centering
    \begin{subfigure}[b]{0.3\textwidth}
        \centering
        \includegraphics[width=\textwidth]{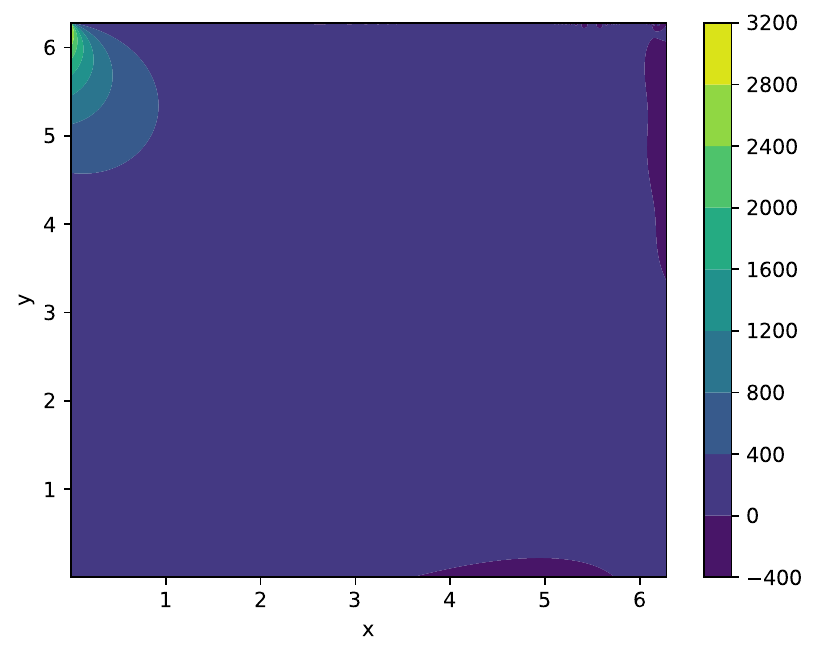}
        \caption{Solution of Temperature Equation with boundary conditions 1}
 \label{fig:heat_1_dist}
    \end{subfigure}
    \hfill
    \begin{subfigure}[b]{0.3\textwidth}
        \centering
        \includegraphics[width=1\textwidth]{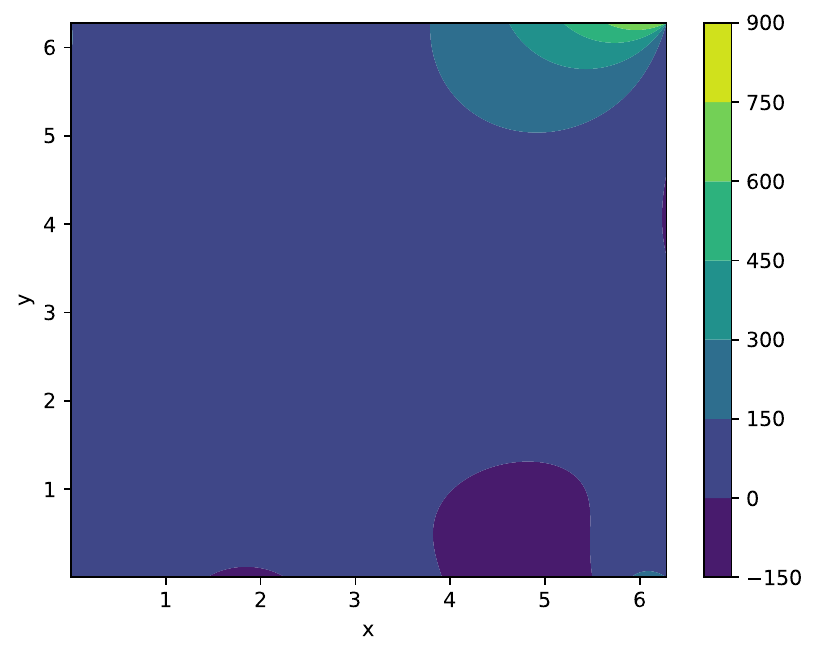}
        \caption{Solution of Temperature Equation with boundary conditions 2}
        \label{fig:heat_2_dist}
    \end{subfigure}
    \hfill
    \begin{subfigure}[b]{0.3\textwidth}
        \centering
        \includegraphics[width=\textwidth]{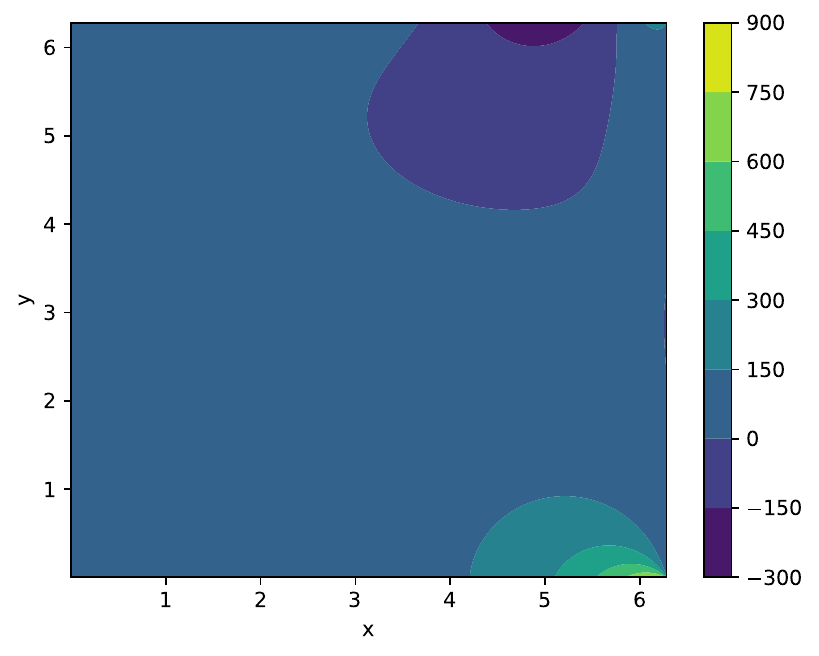}
        \caption{Solution of Temperature Equation with boundary conditions 3}
        \label{fig:heat_3_dist}
    \end{subfigure}
    \caption{The figures depict the solutions for temperature distributions governed by the heat equation, with each figure corresponding to a specific tuple of boundary conditions described in Section \ref{section:experiments_2d_laplace}. It is evident that the region with the highest temperature is relatively small in comparison to the entire domain.}
    \label{fig:heat_dist}
\end{figure}

In this study, we showcase the benchmark optimization outcomes achieved by our proposed PINN-BO algorithm, comparing them with baseline methods for the steady-state temperature optimization task.  The governing PDE dictating the temperature distribution is expressed as: $\nabla^2 T(x, y) = 0$. We explore the heat equation in a domain where $x$ and $y$ lie within the defined range of $[0,2\pi]$. To thoroughly investigate the problem, we consider three distinct boundary conditions for this equation, each contributing to a nuanced understanding of the system: 

 \paragraph{Heat Equation with boundary conditions 1}
 \label{para:heat1}
 \begin{align*}
     T(x, 0) &= 5\sin(y) + \sqrt{1+y} \\
    T(x, 2\pi) &= y\sin\left(3\cos(y) + 2\exp(y)\sin(y)\right) \\
   T(0, y) &= 10\cos(x) + x\exp(\sqrt{x^2 + \sin(x)}) \\
   T(2\pi, y) &= 3\sqrt{\exp(x\exp(-x))}\sin(x) + \cos(3x)\cos(3x)
 \end{align*}
 \paragraph{Heat Equation with boundary conditions 2}
 \label{para:heat2}
\begin{align*}
    T(x, 0) &= \sin(x) \cos(2x) + x^2\sqrt{3x}  + e^{\sin(x)} \\
    T(x, 2\pi) &= e^{\sin(x)} \sqrt{3x} + x^2 \cos(x)  \sin^2(x)   + e^{\cos(x)} \\
    T(0, y) &= \sqrt{2y}  \sin(y) + y^3\cos(2y)   + e^{\cos(y)} \\
    T(2\pi, y) &= \sin(y) \cos(2y) + y^3\sqrt{2y}   + e^{\sin(y)}
\end{align*}
\paragraph{Heat Equation with boundary conditions 3}
\label{para:heat3}
\begin{align*}
    T(x, 0) &= \left(\sin(x) + \cos(2x) \right) \sqrt{3x} + x^2 + e^{\sin(x)} \\
    T(x, 2\pi) &= \left(e^{\sin(x)} + \sqrt{3x} \right) \cos(x) + \left(\sin^2(x) + x^2\right)  e^{\cos(x)}
    \\
    T(0, y) &= \left(\sqrt{2y} + \sin(y)\right)  \left(\cos(2y) + y^3 \right) + e^{\cos(y)}\\
    T(2\pi, y) &= \left(\sin(y) + \cos(2y)\right) \left(\sqrt{2y} + y^3\right) + e^{\sin(y)}
\end{align*}
We utilized py-pde, a Python package designed for solving partial differential equations (PDEs), available at the following GitHub repository: \url{https://github.com/zwicker-group/py-pde}. This tool enabled us to obtain solutions to heat equations at various input points, with specific boundary conditions serving as the input data. In Figure \ref{fig:heat_dist}, the temperature distribution within the defined domain $[0, 2\pi]$ is visualized. It is important to emphasize that the boundary conditions used for benchmarking purposes are unknown to the methods employed. Adhering to the framework of black-box optimization, we assume that solving the PDEs incurs a substantial computational cost. The figure illustrates the spatial distribution of temperature values $T(x,y)$ across domain $[0, 2\pi] \times [0, 2\pi]$, with each subfigure corresponding to one of the aforementioned boundary conditions. 

\begin{figure}[!ht]
    \centering
    \begin{subfigure}[b]{0.3\textwidth}
        \centering
    \includegraphics[width=\textwidth]{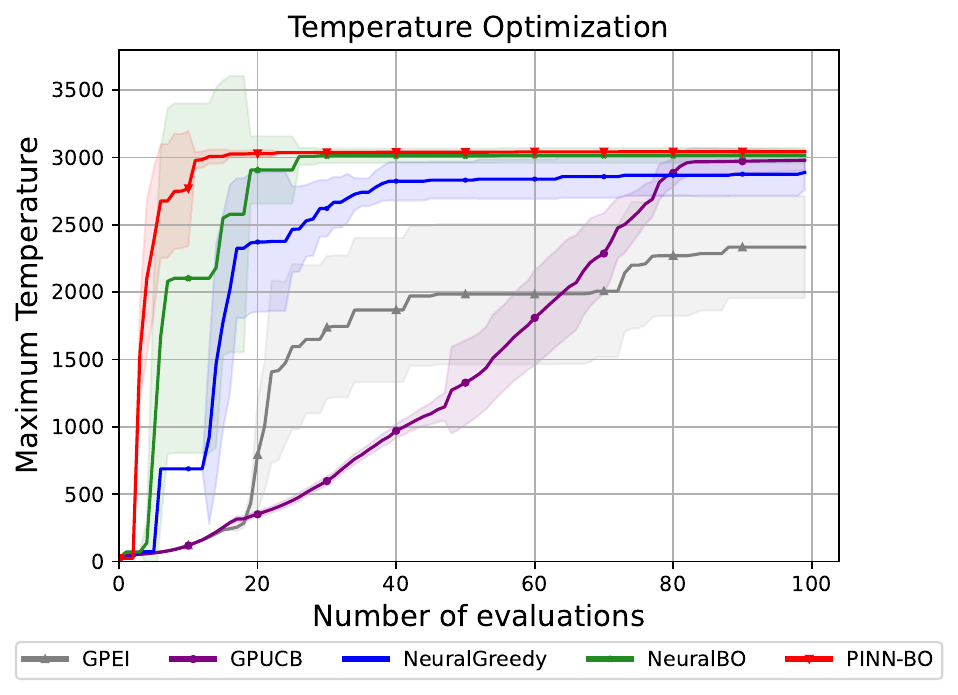}
        \caption{Temperature Optimization in case of boundary conditions 1}
 \label{fig:heat_1_opt}
    \end{subfigure}
    \hfill
    \begin{subfigure}[b]{0.3\textwidth}
        \centering
        \includegraphics[width=1\textwidth]{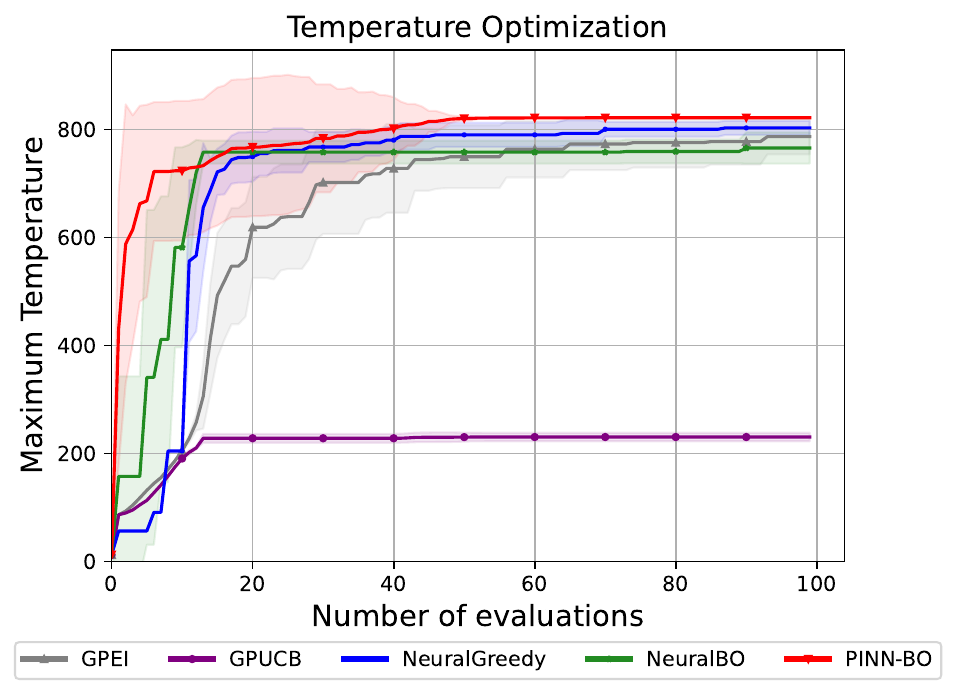}
        \caption{Temperature Optimization in case of boundary conditions 2}
        \label{fig:heat_2_opt}
    \end{subfigure}
    \hfill
    \begin{subfigure}[b]{0.3\textwidth}
        \centering
        \includegraphics[width=\textwidth]{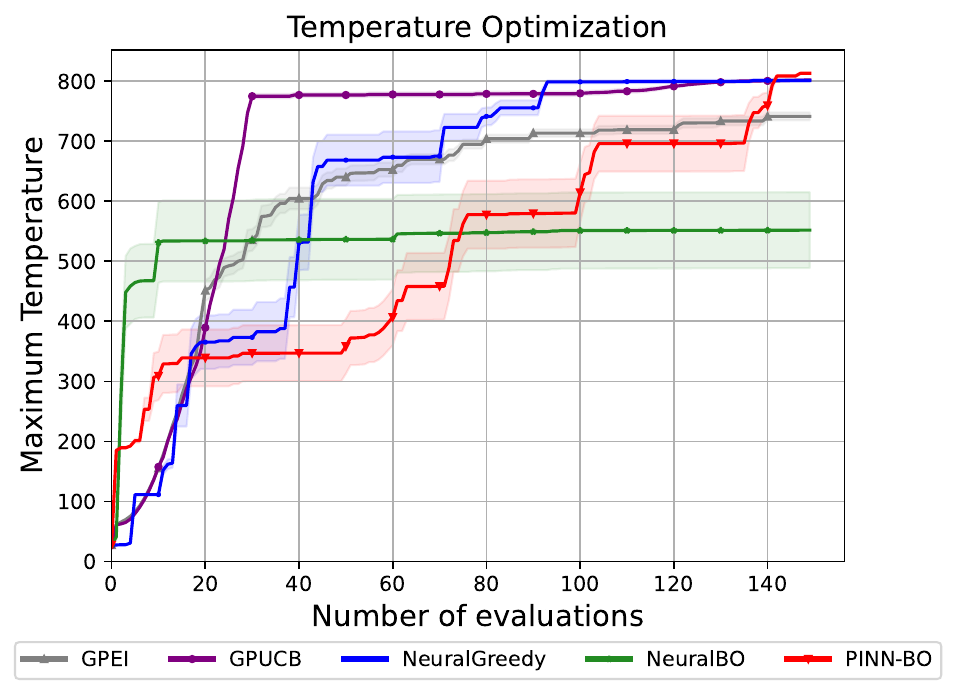}
        \caption{Temperature Optimization in case of boundary conditions 3}
        \label{fig:heat_3_opt}
    \end{subfigure}
    \caption{The figure shows the temperature optimization results of our PINN-BO and other baselines. For all three cases with different positions of maximum temperature, our PINN-BO performs better than all other baselines.}
    \label{fig:heat_opt}
\end{figure}

We conducted temperature optimization by identifying the locations $(x,y)$ where the temperature reaches its maximum. As illustrated in Figure \ref{fig:heat_dist}, the area with high temperatures is relatively small in comparison to the regions with medium or low temperatures. For each baseline, we performed the optimization process 10 times, computing the average results. The comparative outcomes are presented in Figure \ref{fig:heat_opt}. 

\subsubsection{Optimizing Beam Displacement}
\begin{figure}[!ht]
    \centering
    \begin{subfigure}[b]{0.44\textwidth}
        \centering
    \includegraphics[width=\textwidth]{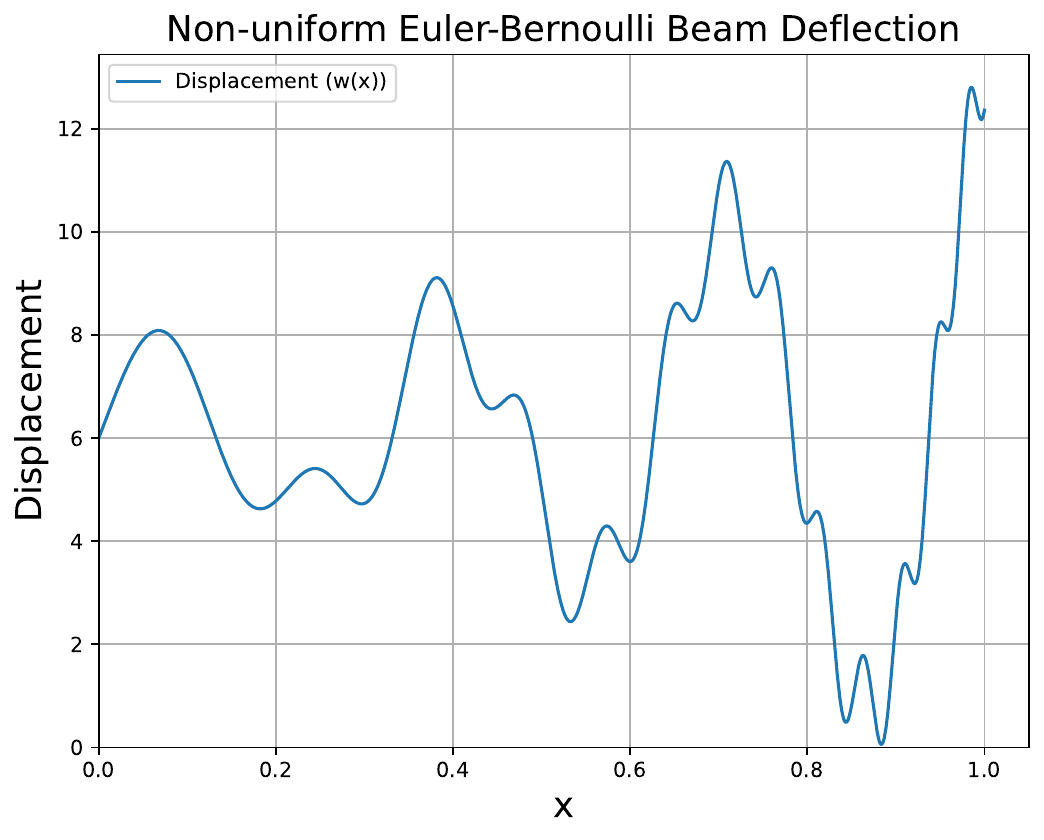}
    \caption{Displacement of non-uniform Euler beam}
     \label{fig:beam_illustration}
    \end{subfigure}
    \hfill
    \begin{subfigure}[b]{0.49\textwidth}
        \centering
        \includegraphics[width=1\textwidth]{figures/BeamDeflection_dim_1.pdf}
        \caption{Displacement Optimization of non-uniform Euler beam}
        \label{fig:beam_opt}
    \end{subfigure}
    \caption{Displacement of a Non-uniform Euler Beam and minimum displacement found by our PINN-BO and the other baselines. The left panel illustrates the natural displacement profile of the non-uniform Euler beam under given loads $q(x)$,  flexural rigidity $EI(x)$, and boundary conditions. The right panel depicts the optimized position on the beam where the displacement is minimized, highlighting the location where the structural response is at its lowest. }
    \label{fig:beam}
\end{figure}

We showcase the benchmark optimization outcomes obtained through our proposed method, PINN-BO, and the baseline approaches, addressing the task of minimizing the deflection of a non-uniform Euler-Bernoulli beam. The governing differential equation describing the behavior of a non-uniform Euler-Bernoulli beam is provided below: 
\begin{equation*}
    \frac{d^2}{dx^2} \left( EI(x) \frac{d^2 w(x)}{dx^2} \right) = q(x),
\end{equation*}
where
$EI(x)$ represents the flexural rigidity of the beam, which can vary with position $x$, and $w(x)$ represents the vertical displacement of the beam at position $x$ and 
$q(x)$ represents the distributed or concentrated load applied to the beam. 
In our implementation, we consider the detailed expression of $EI(x)$ and $q(x)$ as follows:
\begin{align*}
    EI(x) &= \frac{e^x}{\rho(x)},\\
    \rho(x) &= 2.4 x - 64 \pi^{2} e^{4 x} \sin{\left(4 \pi e^{2 x} \right)} - 396 e^{2 x} \sin{\left(20 x \right)} \\
    &+ 80 e^{2 x} \cos{\left(20 x \right)} + 16 \pi e^{2 x} \cos{\left(4 \pi e^{2 x} \right)} + 0.4
\end{align*}
We employed the Finite Difference Method (FDM) to solve the non-uniform Euler-Bernoulli beam equation. It's crucial to note that this step is solely for generating observations at each input point. Despite obtaining this solution, our methods and all baseline techniques continue to treat this solution as a black-box function, accessing observations solely through querying. In Figure \ref{fig:beam_illustration}, the displacement values $w(x)$ for $x \in (0,1)$ are illustrated. The optimization results for both our PINN-BO and the other baseline methods are presented in Figure \ref{fig:beam_opt}.

\section{Proofs of Theoretical Results}
\label{section:proof}
\subsection{Proof of Lemma \ref{lemma:PINN_mean_cov}}
In this section, we present the detailed proof of Lemma \ref{lemma:PINN_mean_cov} in the main paper. Before going to the proof, we repeat Lemma \ref{lemma:PINN_mean_cov} here:
\PinnMeanCov*

\textbf{To prove Lemma \ref{lemma:PINN_mean_cov}, we need the following lemma:}

\begin{sublemma}{(Theorem 4.1 \cite{wang2022and})}
\label{lemma:PINN_GP}
A sufficiently wide physics-informed neural network for modeling the problem defined in Section 4 induces a joint multivariate
Gaussian distribution between the network outputs and its ``derivatives'' after applying the differential operator to this network
\begin{align}
    \begin{bmatrix}
        \mathbf{h}_{\boldsymbol{\theta}} \\
        \mathbf{g}_{\boldsymbol{\theta}} 
\end{bmatrix} \sim N(\mathbf{0}, \mathbf{K}_\mathrm{NTK-PINN}),
\end{align} 
where $\mathbf{K}_\mathrm{NTK-PINN} = \begin{bmatrix}
    \mathbf{K}_{uu} & \mathbf{K}_{ur} \\
    \mathbf{K}_{ru} & \mathbf{K}_{rr}
\end{bmatrix}$ is the NTK matrix of PINN, with
\begin{align}
    (\mathbf{K}_{uu})_{ij} &= \langle \phi(\mathbf{x}_i), \phi(\mathbf{x}_j) \rangle \\ 
    (\mathbf{K}_{ur})_{ij} &= \langle \phi(\mathbf{x}_i), \omega(\mathbf{z}_j) \rangle
    \\
    (\mathbf{K}_{uu})_{ij} &= \langle \omega(\mathbf{z}_i), \omega(\mathbf{z}_j) \rangle
    \\
    \mathbf{K}_{ru} &= \mathbf{K}_{ur}^\top, \\ 
    \mathbf{h}_{\boldsymbol{\theta}} &= [h(\mathbf{x}_1, \boldsymbol{\theta}), \dots, h(\mathbf{x}_t, \boldsymbol{\theta})]^\top
    \\
    \mathbf{g}_{\boldsymbol{\theta}} &= \left[\mathcal{N}[h](\mathbf{z}_1, \boldsymbol{\theta}), \dots, \mathcal{N}[h](\mathbf{z}_{N_r}, \boldsymbol{\theta}) \right]^\top
\end{align}
where $\mathbf{x}_i, \mathbf{x}_j \in \mathcal{D}_t$ and $\mathbf{z}_i, \mathbf{z}_j \in \mathcal{R}$ are two arbitrary points belonging to the set $\mathcal{R}$ defined in Algorithm \ref{alg:PINN-BO} and $\mathcal{N}[h]$ denotes a differential operator of the neural network $h(\cdot, \boldsymbol{\theta})$ with respect to the input $\mathbf{x}$. 
\end{sublemma}
\begin{subcorollary}
\label{corollary:PINN_GP_func}
    The unknown reward function values $f_{1:t}$ and the PDE observations $g_{1:N_r}$ are jointly Gaussian with an initial prior distribution with zero mean and the covariance $\nu_t^2 \mathbf{K}_\mathrm{NTK-PINN}$, where $\nu_t$ is the exploration coefficient introduced in Section \ref{section:theoretical_analysis}. 
    \begin{align}
    \begin{bmatrix}
        f_{1:t} \\
        g_{1:N_r}
\end{bmatrix} \sim N(\mathbf{0}, \nu_t^2 \mathbf{K}_\mathrm{NTK-PINN}),
\end{align} 
\end{subcorollary}

\begin{proof}[Proof of Corollary  \ref{corollary:PINN_GP_func}]
The algorithm updates the neural network by minimizing the loss function: \begin{equation}
    \mathcal{L}(t) = \sum^{t-1}_{i=1} [y_i - \nu_t h(\mathbf{x}_i; \boldsymbol{\theta}_{t-1})]^2 + \sum^{N_r}_{j=1}[u_j - \nu_t \mathcal{N}[h](\mathbf{z}_j; \boldsymbol{\theta}_{t-1})]^2
\end{equation} 
As the function prediction and its ``derivatives" prediction of each observation $\mathbf{x}_i$ and $\mathbf{z}_j$ at iteration $t$ is modeled by $\nu_t h(\mathbf{x}_i; \boldsymbol{\theta}_{t-1})$ and $\nu_t \mathcal{N}[h](\mathbf{z}_j; \boldsymbol{\theta}_{t-1})$, it is clear to see, from Lemma \ref{lemma:PINN_GP}, that the function values of the unknown function $f$ can be assumed to follow a joint Gaussian prior with zero means and covariance matrix $\nu_t^2 \mathbf{K}_\mathrm{NTK-PINN}$. A similar argument can be applied to the function $g$, where $g(\cdot) = \mathcal{N}[f] (\cdot)$ with $\mathcal{N}[f]$ is a differential operator. 
\end{proof}

\begin{proof} [Proof of Lemma \ref{lemma:PINN_mean_cov}]
From Corollary \ref{corollary:PINN_GP_func}, the priors for values of both $f$ and $g$ follow a joint Gaussian distribution with kernel $\mathbf{K}_\mathrm{NTK-PINN}$. Let $\mathbf{K_x}$ be the NTK matrix between point $\mathbf{x}$ and all training data: $\mathbf{K}_x = \begin{bmatrix}
        \Sigma_a & \Sigma_b \\
        \Sigma_c & \Sigma_d
    \end{bmatrix} =\begin{bmatrix}
         \Phi_t \phi(\mathbf{x}) & 
         \Phi_t \omega(\mathbf{x}) \\
         \Omega_r\phi(\mathbf{x}) &
          \Omega_r  \omega(\mathbf{x})  \end{bmatrix}$. 
Then the the posterior of $f$ and $g$ evaluated at an input $\mathbf{x}$ will be a Gaussian distribution with mean and variance functions:
\paragraph{Mean function:}
\begin{align}
    \begin{bmatrix}
        \mu_t^f(\mathbf{x})\\
        \mu_t^g(\mathbf{x}) 
    \end{bmatrix} &= \mathbf{K}_\mathbf{x} ^\top \mathbf{\widehat{K}}_\mathrm{PINN}^{-1} \begin{bmatrix}
        \mathbf{Y}_t \\
        \mathbf{U}_r
    \end{bmatrix}\\
    &= \begin{bmatrix}
        \Sigma_a^\top & \Sigma_c^\top \\
        \Sigma_b^\top & \Sigma_d^\top
    \end{bmatrix} \begin{bmatrix}
        \widetilde{\mathbf{A}} & \widetilde{\mathbf{B}} \\
    \widetilde{\mathbf{C}} & \widetilde{\mathbf{D}}
    \end{bmatrix} \begin{bmatrix}
        \mathbf{Y}_t \\
        \mathbf{U}_r
    \end{bmatrix}\\
    &= \begin{bmatrix}
        \phi(\mathbf{x})^\top \Phi_t^\top &  \phi(\mathbf{x})^\top \Omega_r ^\top \\
        \omega(\mathbf{x})^\top \Phi_t^\top  & \omega(\mathbf{x})^\top \Omega_r ^\top
    \end{bmatrix} \begin{bmatrix}
        \widetilde{\mathbf{A}} & \widetilde{\mathbf{B}} \\
    \widetilde{\mathbf{C}} & \widetilde{\mathbf{D}}
    \end{bmatrix} \begin{bmatrix}
        \mathbf{Y}_t \\
        \mathbf{U}_r
    \end{bmatrix} \\
    & = \begin{bmatrix}
         \phi(\mathbf{x})^\top \Phi_t^\top \widetilde{\mathbf{A}}\mathbf{Y}_t +  \phi(\mathbf{x})^\top \Omega_r ^\top  \widetilde{\mathbf{C}}\mathbf{Y}_t  + \phi(\mathbf{x})^\top \Phi_t^\top \widetilde{\mathbf{B}} \mathbf{U}_r   +  \phi(\mathbf{x})^\top \Omega_r ^\top  \widetilde{\mathbf{D}} \mathbf{U}_r \\
         \omega(\mathbf{x})^\top \Phi_t^\top \widetilde{\mathbf{A}} \mathbf{Y}_t +  \omega(\mathbf{x})^\top \Omega_r ^\top  \widetilde{\mathbf{C}} \mathbf{Y}_t  +\omega(\mathbf{x})^\top \Phi_t^\top \widetilde{\mathbf{B}} \mathbf{U}_r  +  \omega(\mathbf{x})^\top \Omega_r ^\top  \widetilde{\mathbf{D}} \mathbf{U}_r
    \end{bmatrix}
    \\
         &=\phi(\mathbf{x})^\top \boldsymbol{\xi}_t^\top \mathbf{\widehat{K}}_\mathrm{PINN}^{-1} \begin{bmatrix}
         \mathbf{Y}_t \\
         \mathbf{U}_r\end{bmatrix}
\end{align}

\paragraph{Variance function:}
Let $\mathbf{K}_{\mathbf{xx}} = \begin{bmatrix}
\langle \phi(\mathbf{x}), \phi(\mathbf{x}) \rangle & \langle \phi(\mathbf{x}), \omega(\mathbf{x}) \rangle \\
\langle \omega(\mathbf{x}), \phi(\mathbf{x}) \rangle & \langle \omega(\mathbf{x}), \omega(\mathbf{x}) \rangle
\end{bmatrix}$, then we have the posterior covariance matrix of $f$ and $g$ at input $\mathbf{x}$ is:
\begin{align}
    \begin{bmatrix}
    \mathrm{Cov}_f(\mathbf{x}) & \mathrm{Cov}_{fg}(\mathbf{x}) \\
    \mathrm{Cov}_{gf}(\mathbf{x}) & \mathrm{Cov}_g(\mathbf{x})
\end{bmatrix}  &= \nu_t^2 \left(\mathbf{K}_{\mathbf{xx}} - \mathbf{K_x}^\top \mathbf{\widehat{K}}_\mathrm{PINN}^{-1} \mathbf{K_x} \right) \\
&= \nu_t^2   \begin{bmatrix}
\langle \phi(\mathbf{x}), \phi(\mathbf{x}) \rangle & \langle \phi(\mathbf{x}), \omega(\mathbf{x}) \rangle \\
\langle \omega(\mathbf{x}), \phi(\mathbf{x}) \rangle & \langle \omega(\mathbf{x}), \omega(\mathbf{x}) \rangle
\end{bmatrix} - \nu_t^2
\begin{bmatrix}
        \Sigma_a^\top& \Sigma_c^\top \\
        \Sigma_b^\top & \Sigma_d^\top
\end{bmatrix} \begin{bmatrix}
    \widetilde{\mathbf{A}} & \widetilde{\mathbf{B}} \\
    \widetilde{\mathbf{C}} & \widetilde{\mathbf{D}}
        \end{bmatrix}  \begin{bmatrix}
                            \Sigma_a & \Sigma_b \\
                            \Sigma_c & \Sigma_d
                        \end{bmatrix}  \\
\end{align}
Therefore, 
\begin{align}
\mathrm{Cov}_f(\mathbf{x}) &= \nu_t^2 \langle \phi(\mathbf{x}),  \phi(\mathbf{x}) \rangle - \nu_t^2(\Sigma_a^\top \widetilde{\mathbf{A}}\Sigma_a + \Sigma_c^\top \widetilde{\mathbf{C}}\Sigma_c + \Sigma_a^\top \widetilde{\mathbf{B}}\Sigma_a + \Sigma_c^\top \widetilde{\mathbf{D}}\Sigma_c) \\
        &= \nu_t^2 \langle \phi(\mathbf{x}),  \phi(\mathbf{x}) \rangle - \nu_t^2 \begin{bmatrix}
            \Sigma_a^\top  & \Sigma_c^\top
        \end{bmatrix}
        \begin{bmatrix}
    \widetilde{\mathbf{A}} & \widetilde{\mathbf{B}} \\
    \widetilde{\mathbf{C}} & \widetilde{\mathbf{D}}
        \end{bmatrix} \begin{bmatrix}
            \Sigma_a \\ \Sigma_c
        \end{bmatrix}  \\
        &= \nu_t^2 \langle \phi(\mathbf{x}),  \phi(\mathbf{x}) \rangle - \nu_t^2\begin{bmatrix}
            \phi(\mathbf{x})^\top \Phi_t^\top & \phi(\mathbf{x})^\top \Omega_r^\top
        \end{bmatrix}  \begin{bmatrix}
    \widetilde{\mathbf{A}} & \widetilde{\mathbf{B}} \\
    \widetilde{\mathbf{C}} & \widetilde{\mathbf{D}}
        \end{bmatrix} \begin{bmatrix} \Phi_t \phi(\mathbf{x}) \\ \Omega_r \phi(\mathbf{x})  \end{bmatrix}  \\
        &= \nu_t^2 \langle \phi(\mathbf{x}),  \phi(\mathbf{x}) \rangle - \nu_t^2\phi(\mathbf{x})^\top \boldsymbol{\xi}_t^\top \mathbf{\widehat{K}}_\mathrm{PINN}^{-1} \boldsymbol{\xi}_t \phi(\mathbf{x})\\
        &= \nu_t^2 (\sigma_t^f)^2(\mathbf{x})
\end{align}
\end{proof}
\subsection{Proof of Lemma \ref{lemma:interaction_information_formula}}
\InteractionInformation*
\textbf{To prove Lemma \ref{lemma:interaction_information_formula}, we need to prove the following technical lemma:}
\begin{sublemma}    \label{lemma:det_division}
    Let $\mathbf{u} \in \mathrm{R}^{n \times p}$ and $\mathbf{K} \in \mathrm{R}^{p \times p}$ is a positive semi-definite matrix and $p \ge n$. Then 
    \begin{align}
    \frac{\det [\mathbf{u}\left(\mathbf{K}(\mathbf{u}^\top\mathbf{u}+\mathbf{I})^{-1} + \mathbf{I}\right)^{-1}\mathbf{u}^\top]}{[\mathbf{u}\left(\mathbf{K} + \mathbf{I}\right)^{-1}\mathbf{u}^\top]} =  \frac{\det \left(\mathbf{K}(\mathbf{u}^\top\mathbf{u}+\mathbf{I})^{-1} + \mathbf{I}\right)^{-1}}{\det \left(\mathbf{K} + \mathbf{I}\right)^{-1}}   
    \end{align}
\end{sublemma}
\begin{proof}[Proof of Lemma \ref{lemma:det_division}]
We start the proof by gradually calculating denominator and numerator

\paragraph{Denominator}

\begin{align}
\det[\mathbf{u} (\mathbf{I} + \mathbf{K})^{-1} \mathbf{u}^\top] &= \det [\mathbf{u} \left[ (\mathbf{I} - \mathbf{K}(\mathbf{I}+\mathbf{K})^{-1}\right]\mathbf{u}^\top]\\
&=\det [\mathbf{u} \mathbf{u}^\top - \mathbf{u} \mathbf{K}(\mathbf{I}+\mathbf{K})^{-1} \mathbf{u}^\top]  \\
&= \det \left[  (\mathbf{u} \mathbf{u}^\top) \left(\mathbf{K}(\mathbf{I}+\mathbf{K})^{-1}\right) \left( (\mathbf{I} + \mathbf{K}) \mathbf{K}^{-1} - \mathbf{u}^\top (\mathbf{u} \mathbf{u}^\top)^{-1} \mathbf{u}\right) \right] \\
&= \det (\mathbf{u} \mathbf{u}^\top) \det \left(\mathbf{K}(\mathbf{I}+\mathbf{K})^{-1}\right) \det \left( (\mathbf{I} + \mathbf{K}) \mathbf{K}^{-1} - \mathbf{u}^\top (\mathbf{u} \mathbf{u}^\top)^{-1} \mathbf{u}\right) \\
&= \det (\mathbf{u} \mathbf{u}^\top) \det \left((\mathbf{I}+\mathbf{K})^{-1}\right) \det \mathbf{K} \det \left( (\mathbf{I} + \mathbf{K}) \mathbf{K}^{-1} - \mathbf{u}^\top (\mathbf{u} \mathbf{u}^\top)^{-1} \mathbf{u}\right) \\
& = \det (\mathbf{u} \mathbf{u}^\top) \det \left((\mathbf{I}+\mathbf{K})^{-1}\right) \det \left( \mathbf{K}(\mathbf{I} + \mathbf{K}) \mathbf{K}^{-1} - \mathbf{K}\mathbf{u}^\top (\mathbf{u} \mathbf{u}^\top)^{-1} \mathbf{u}\right) \\
\label{Eqn:u_K_inv_uT}
& = \det (\mathbf{u} \mathbf{u}^\top) \det \left((\mathbf{I}+\mathbf{K})^{-1}\right) \det \left(\mathbf{I} + \mathbf{K} - \mathbf{K}\mathbf{u}^\top (\mathbf{u} \mathbf{u}^\top)^{-1} \mathbf{u}\right)
\end{align}
The first equation utilizes the  Woodburry matrix inversion formula while the third equation uses generalized matrix determinant lemma \footnote{Suppose $\mathbf{A}$ is an invertible $n$-by-$n$ matrix and $\mathbf{U}, \mathbf{V}$ are $n$-by-$m$ matrices, $m \le n$. Then $\det(\mathbf{A} + \mathbf{U}\mathbf{W}\mathbf{V}^\top) = \det(\mathbf{A}) \det(\mathbf{W})\det(\mathbf{W}^{-1} + \mathbf{V}^\top \mathbf{A}^{-1} \mathbf{U})$.}.
\paragraph{Numerator}
Let $\widetilde{\mathbf{K}} = \mathbf{K} (\mathbf{u}^\top \mathbf{u} + \mathbf{I})^{-1}$, then we have
\begin{align}
    \det [\mathbf{u}\left(\mathbf{K}(\mathbf{u}^\top\mathbf{u}+\mathbf{I})^{-1} + \mathbf{I}\right)^{-1}\mathbf{u}^\top] &= \det [\mathbf{u}\left( \mathbf{I}+\widetilde{\mathbf{K}} \right)^{-1}\mathbf{u}^\top] \\
    & = \det (\mathbf{u} \mathbf{u}^\top) \det \left((\mathbf{I}+\widetilde{\mathbf{K}})^{-1}\right) \det \left( \mathbf{I} + \widetilde{\mathbf{K}} - \widetilde{\mathbf{K}}\mathbf{u}^\top (\mathbf{u} \mathbf{u}^\top)^{-1} \mathbf{u}\right),
\end{align}
where we use the result at line  (\ref{Eqn:u_K_inv_uT}) and replace $\mathbf{K}$ by $\widetilde{\mathbf{K}}$. We also have: 
\begin{align}
    &\det \left( (\mathbf{I} + \widetilde{\mathbf{K}}) - \widetilde{\mathbf{K}}\mathbf{u}^\top (\mathbf{u} \mathbf{u}^\top)^{-1} \mathbf{u}\right) \\
    = &\det \left( \mathbf{I} + \mathbf{K} (\mathbf{u}^\top \mathbf{u} + \mathbf{I})^{-1} - \mathbf{K} (\mathbf{u}^\top \mathbf{u} + \mathbf{I})^{-1}\mathbf{u}^\top (\mathbf{u} \mathbf{u}^\top)^{-1} \mathbf{u}\right) \\
    = &  \det \left[ \mathbf{I} + \mathbf{K} \left( \mathbf{I} - \mathbf{u}^\top \mathbf{u} (\mathbf{u}^\top \mathbf{u} + \mathbf{I})^{-1} \right) - \mathbf{K} (\mathbf{u}^\top \mathbf{u} + \mathbf{I})^{-1}\mathbf{u}^\top (\mathbf{u} \mathbf{u}^\top)^{-1} \mathbf{u}\right] \\
    = &  \det \left[ \mathbf{I} + \mathbf{K} - \mathbf{K} \mathbf{u}^\top \mathbf{u} (\mathbf{u}^\top \mathbf{u} + \mathbf{I})^{-1}  - \mathbf{K} (\mathbf{u}^\top \mathbf{u} + \mathbf{I})^{-1}\mathbf{u}^\top (\mathbf{u} \mathbf{u}^\top)^{-1} \mathbf{u}\right] \\
    = &  \det \left[ \mathbf{I} + \mathbf{K} - \mathbf{K} \mathbf{u}^\top (\mathbf{u} \mathbf{u}^\top + \mathbf{I})^{-1}\mathbf{u}  - \mathbf{K} \mathbf{u}^\top ( \mathbf{u}\mathbf{u}^\top + \mathbf{I})^{-1} (\mathbf{u} \mathbf{u}^\top)^{-1} \mathbf{u}\right] \\ 
    = &  \det \left[ \mathbf{I} + \mathbf{K} - \mathbf{K} \mathbf{u}^\top (\mathbf{u} \mathbf{u}^\top + \mathbf{I})^{-1} \left( \mathbf{I} + (\mathbf{u} \mathbf{u}^\top)^{-1} \right)  \mathbf{u} \right] \\ 
    = &  \det \left( \mathbf{I} + \mathbf{K} - \mathbf{K} \mathbf{u}^\top (\mathbf{u} \mathbf{u}^\top)^{-1}  \mathbf{u} \right)
\end{align}
Therefore, we have the final expression of the numerator  
\begin{align}
    \det \left( (\mathbf{I} + \widetilde{\mathbf{K}}) - \widetilde{\mathbf{K}}\mathbf{u}^\top (\mathbf{u} \mathbf{u}^\top)^{-1} \mathbf{u}\right) = \det (\mathbf{u} \mathbf{u}^\top) \det \left((\mathbf{I}+\widetilde{\mathbf{K}})^{-1}\right) \det \left( \mathbf{I} + \mathbf{K} - \mathbf{K} \mathbf{u}^\top (\mathbf{u} \mathbf{u}^\top)^{-1}  \mathbf{u} \right)
\end{align}
Using derived numerator and denominator, we have 
\begin{align}
    \frac{\det [\mathbf{u}\left(\mathbf{K}(\mathbf{u}^\top\mathbf{u}+\mathbf{I})^{-1} + \mathbf{I}\right)^{-1}\mathbf{u}^\top]}{[\mathbf{u}\left(\mathbf{K} + \mathbf{I}\right)^{-1}\mathbf{u}^\top]} &= \frac{\det (\mathbf{u} \mathbf{u}^\top) \det \left((\mathbf{I}+\widetilde{\mathbf{K}})^{-1}\right) \det \left( \mathbf{I} + \mathbf{K} - \mathbf{K} \mathbf{u}^\top (\mathbf{u} \mathbf{u}^\top)^{-1}  \mathbf{u} \right)}{\det (\mathbf{u} \mathbf{u}^\top) \det \left((\mathbf{I}+\mathbf{K})^{-1}\right) \det \left(\mathbf{I} + \mathbf{K} - \mathbf{K}\mathbf{u}^\top (\mathbf{u} \mathbf{u}^\top)^{-1} \mathbf{u}\right)}\\
    &=\frac{\det (\mathbf{I}+\widetilde{\mathbf{K}})^{-1}}{\det (\mathbf{I}+\mathbf{K})^{-1}} \\
    &=\frac{\det \left(\mathbf{K}(\mathbf{u}^\top\mathbf{u}+\mathbf{I})^{-1} + \mathbf{I}\right)^{-1}}{\det \left(\mathbf{K} + \mathbf{I}\right)^{-1}}   
    \end{align} 
\end{proof}

\begin{subcorollary}
\label{corollary:det_divison_corollary}
Let $\mathbf{u} \in \mathrm{R}^{n \times p}$ and $\mathbf{K} \in \mathrm{R}^{p \times p}$ is a positive semi-definite matrix and $p \ge n$. Then 
\begin{align}
    \frac{\det[\mathbf{u}(\mathbf{K}+\mathbf{I})^{-1} \mathbf{u}^\top]}{\det [\mathbf{u}(\mathbf{K}+ \mathbf{u}^\top \mathbf{u} + \mathbf{I})^{-1} \mathbf{u}^\top]} = \frac{\det (\mathbf{K}+\mathbf{I})^{-1}}{(\mathbf{K}+\mathbf{u}^\top \mathbf{u}+\mathbf{I})^{-1}}
\end{align}
\end{subcorollary}
\begin{proof} [Proof of Corollary \ref{corollary:det_divison_corollary}]
    \begin{align}
        \frac{\det[\mathbf{u}(\mathbf{K}+\mathbf{I})^{-1} \mathbf{u}^\top]}{\det [\mathbf{u}(\mathbf{K}+ \mathbf{u}^\top \mathbf{u} + \mathbf{I})^{-1} \mathbf{u}^\top]} &= \frac{\det[\mathbf{u}(\mathbf{K}+\mathbf{I})^{-1} \mathbf{u}^\top]}{\det [\mathbf{u} (\mathbf{u}^\top \mathbf{u}+\mathbf{I})^{-1}(\mathbf{K} (\mathbf{u}^\top \mathbf{u} + \mathbf{I})^{-1} + \mathbf{I})^{-1} \mathbf{u}^\top]}
        \\
        &=\frac{\det[\mathbf{u}(\mathbf{K}+\mathbf{I})^{-1} \mathbf{u}^\top]}{\det [ (\mathbf{u} \mathbf{u}^\top+\mathbf{I})^{-1}\mathbf{u} (\mathbf{K} (\mathbf{u}^\top \mathbf{u} + \mathbf{I})^{-1} + \mathbf{I})^{-1} \mathbf{u}^\top]}
        \\
        &= \frac{1}{\det (\mathbf{u} \mathbf{u}^\top+\mathbf{I})^{-1}} \frac{\det[\mathbf{u}(\mathbf{K}+\mathbf{I})^{-1} \mathbf{u}^\top]}{\det [\mathbf{u} (\mathbf{K} (\mathbf{u}^\top \mathbf{u} + \mathbf{I})^{-1} + \mathbf{I})^{-1} \mathbf{u}^\top]}
        \\
        &=\frac{1}{\det (\mathbf{u} ^\top \mathbf{u}+\mathbf{I})^{-1}} \frac{\det[(\mathbf{K}+\mathbf{I})^{-1}]}{\det [ (\mathbf{K} (\mathbf{u}^\top \mathbf{u} + \mathbf{I})^{-1} + \mathbf{I})^{-1}]}
        \\
        &= \frac{\det (\mathbf{K}+\mathbf{I})^{-1}}{(\mathbf{K}+\mathbf{u}^\top \mathbf{u}+\mathbf{I})^{-1}},
    \end{align}
The second equation uses the matrix inversion identity of two non-singular matrices $\mathbf{A}$ and $\mathbf{B}$, i.e., $(\mathbf{A}\mathbf{B})^{-1} = \mathbf{B}^{-1} \mathbf{A}^{-1}$ while the fourth equation directly utilizes Lemma \ref{lemma:det_division}.
\end{proof}
\begin{proof}[Proof of Lemma \ref{lemma:interaction_information_formula}]
By definition, we have $I (f; \mathbf{Y}_t; \mathbf{U}_r) = I(f; \mathbf{Y}_t) - I (f; \mathbf{Y}_t \rvert \mathbf{U}_r)$. We start by proof by calculating $I (f; \mathbf{Y}_t \rvert \mathbf{U}_r)$.

By the properties of GPs, given a set of sampling points $\mathcal{D}_t\subset \mathcal{D}$, we have that $f, \mathbf{Y}_t, \mathbf{U}_r$ are jointly Gaussian:
\begin{align}
\label{Eqn:joint_gauss_3_variables}
\left(
\begin{aligned}
    f\\
    \mathbf{Y}_t\\
    \mathbf{U}_r
\end{aligned}
\right) \sim \mathcal{N} \left(\mathbf{0}, \nu_t^2 \begin{bmatrix}
    \mathbf{K}_{uu} & \mathbf{K}_{uu} & \mathbf{K}_{ur} \\
    \mathbf{K}_{uu} & \mathbf{K}_{uu} + \lambda_1 \mathbf{I} & \mathbf{K}_{ur} \\
    \mathbf{K}_{ru} & \mathbf{K}_{uu} & \mathbf{K}_{rr} + \lambda_2 \mathbf{I}
\end{bmatrix} \right)    
\end{align}

Then, we have
\begin{align}
    \textup{Cov}(f \rvert \mathbf{U}_r) &=  \nu_t^2 \left[ \mathbf{K}_{uu} - \mathbf{K}_{ur}(\mathbf{K}_{rr}+ \lambda_2\mathbf{I})^{-1} \mathbf{K}_{ru} \right]
        \\
        &= \nu_t^2 \left[\Phi_t \Phi_t^\top - \Phi_t\Omega_r^\top (\Omega_r\Omega_r^\top + \lambda_2\mathbf{I})^{-1} \Omega_r\Phi_t^\top \right]
        \\
        &= \nu_t^2 \left[\Phi_t\Phi_t^\top - \Phi_t\left[\mathbf{I} - \lambda_2(\Omega_r\Omega_r^\top + \lambda_2\mathbf{I})^{-1}\right] \Phi_t^\top \right] 
        \\
        &= \nu_t^2 \lambda_2 \Phi_t (\Omega_r^\top \Omega_r + \lambda_2\mathbf{I})^{-1} \Phi_t^\top
        \\ \label{Eqn:covar_f_condition_on_u}
        &= \nu_t^2 \Phi_t \left(\frac{\Omega_r^\top \Omega_r}{\lambda_2} + \mathbf{I}\right)^{-1} \Phi_t^\top,
\end{align}
and 
\begin{align}
\label{Eqn:covar_f_condition_on_y_and_u}
        \textup{Cov}(f \rvert \mathbf{Y}_t; \mathbf{U}_r) &= \nu_t^2 \left(\mathbf{K}_{uu} - \begin{bmatrix}
            \mathbf{K}_{uu} & \mathbf{K}_{ur}
        \end{bmatrix} \begin{bmatrix}
            \mathbf{K}_{uu} + \lambda_1\mathbf{I} & \mathbf{K}_{ur} \\
            \mathbf{K}_{ru} & \mathbf{K}_{rr} + \lambda_2 \mathbf{I}
        \end{bmatrix}^{-1} \begin{bmatrix}
            \mathbf{K}_{uu} \\
            \mathbf{K}_{ru}
        \end{bmatrix} \right) \\
        & = \nu_t^2 (\Phi_t \Phi_t^\top - \Phi_t \boldsymbol{\xi}_t^\top \mathbf{\widehat{K}}_\mathrm{PINN}^{-1} \boldsymbol{\xi}_t \Phi_t^\top) 
\end{align}
Let $\mathbf{V} = \boldsymbol{\xi}_t^\top \mathbf{\widehat{K}}_\mathrm{PINN}^{-1} \boldsymbol{\xi}_t$, now we need to calculate $\mathbf{V}$. We have
    \begin{align}
            \mathbf{V} &= \boldsymbol{\boldsymbol{\xi}_t}^\top \mathbf{\widehat{K}}_\mathrm{PINN}^{-1} \boldsymbol{\boldsymbol{\xi}_t} \\
            &=\Phi_t^\top \widetilde{\mathbf{A}}\Phi_t + \Omega_r^\top \widetilde{\mathbf{C}}\Phi_t + \Phi_t^\top \widetilde{\mathbf{B}}\Omega_r + \Omega_r^\top \widetilde{\mathbf{D}}\Omega_r \\
            & = \Phi_t^\top (\mathbf{P}^{-1} - \mathbf{P}^{-1}\mathbf{Q}\widetilde{\mathbf{C}})\Phi_t + \Omega_r^\top \widetilde{\mathbf{C}}\Phi_t - \Phi_t^\top \mathbf{P}^{-1}\mathbf{Q}\Omega_r + \Omega_r^\top \widetilde{\mathbf{D}}\Omega_r \\
            & = \Phi_t^\top \mathbf{P}^{-1}\Phi_t - \Phi_t^\top\mathbf{P}^{-1}\mathbf{Q}\widetilde{\mathbf{C}}\Phi_t + \Omega_r^\top \widetilde{\mathbf{C}}\Phi_t - \Phi_t^\top \mathbf{P}^{-1}\mathbf{Q}\Omega_r + \Omega_r^\top \widetilde{\mathbf{D}}\Omega_r \\
            & = \Phi_t^\top \mathbf{P}^{-1}\Phi_t + (\underbrace{\Omega_r - \Phi_t^\top\mathbf{P}^{-1}\mathbf{Q}}_{U_1})(\widetilde{\mathbf{C}}\Phi_t + \widetilde{\mathbf{D}}\Omega_r) \\
            & = \Phi_t^\top \mathbf{P}^{-1}\Phi_t + (\underbrace{\Omega_r - \Phi_t^\top\mathbf{P}^{-1}\mathbf{Q}}_{V_1})(\underbrace{\widetilde{\mathbf{C}}\Phi_t + \widetilde{\mathbf{D}}\Omega_r)}_{V_2} \\
            & = \Phi_t^\top (\Phi_t\Phi_t^\top+\lambda_1 \mathbf{I})^{-1} \Phi_t + V_1V_2\\
            & = (\Phi_t^\top\Phi_t+\lambda_1 \mathbf{I})^{-1} \Phi_t^\top\Phi_t + V_1V_2
    \end{align}
    where 
    \begin{align}
            \label{terms:pqrs}
            \begin{bmatrix}
            \mathbf{P} & \mathbf{Q} \\
            \mathbf{R} & \mathbf{S}
            \end{bmatrix} & = \begin{bmatrix}
            \mathbf{K}_{uu} + \lambda_1\mathbf{I} & \mathbf{K}_{ur}  \\
            \mathbf{K}_{ru}  & \mathbf{K}_{rr} + \lambda_2\mathbf{I}
            \end{bmatrix} = \begin{bmatrix}
            \Phi_t\Phi_t^\top + \lambda_1\mathbf{I} & \Phi_t \Omega_r^\top \\
            \Omega_r\Phi_t^\top & \Omega_r\Omega_r^\top + \lambda_2 \mathbf{I}\end{bmatrix} \\
            \text{and} \begin{bmatrix}
    \widetilde{\mathbf{A}} & \widetilde{\mathbf{B}} \\
    \widetilde{\mathbf{C}} & \widetilde{\mathbf{D}} 
        \end{bmatrix} &=\begin{bmatrix}
            \mathbf{P} & \mathbf{Q} \\
            \mathbf{R} & \mathbf{S}
            \end{bmatrix} ^{-1}
    \end{align}
    The second equality applied the formula of block matrix inversion \footnote{The inversion of matrix $\mathbf{K} = \begin{bmatrix}
        \mathbf{P} & \mathbf{Q} \\ \mathbf{R} & \mathbf{S} 
    \end{bmatrix}$ is given as $\mathbf{K}^{-1} = \begin{bmatrix}
        \mathbf{P}^{-1} + \mathbf{P}^{-1} \mathbf{QMR}\mathbf{P}^{-1} & \mathbf{P}^{-1}\mathbf{QM} \\ -\mathbf{MR}\mathbf{P}^{-1} & \mathbf{M} 
    \end{bmatrix}$ with $\mathbf{M} = (\mathbf{S} - \mathbf{R}\mathbf{P}^{-1}\mathbf{Q})^{-1}$.}, 
    while the last equality used push-through identity. Next, we have
    \begin{align}
            \mathbf{M} &= (\mathbf{S} - \mathbf{R}\mathbf{P}^{-1}\mathbf{Q})^{-1} \\
            &= \left[\Omega_r \Omega_r^\top  + \lambda_2\mathbf{I} - \Omega_r \Phi_t^\top (\Phi_t \Phi_t ^\top+ \lambda_1\mathbf{I})^{-1} \Phi_t \Omega_r^\top)\right]^{-1} \\
            &= \left[\Omega_r \Omega_r^\top  + \lambda_2\mathbf{I} - \Omega_r (\Phi_t^\top \Phi_t+ \lambda_1\mathbf{I})^{-1} \Phi_t^\top \Phi_t \right] \Omega_r^\top)^{-1} \\ 
            &= \left[\Omega_r \Omega_r^\top  + \lambda_2\mathbf{I} - \Omega_r \left[ \mathbf{I} - \lambda_1(\Phi_t^\top \Phi_t+ \lambda_1\mathbf{I})^{-1} \right] \Omega_r^\top)\right]^{-1} \\ 
            &= \left[\lambda_2\mathbf{I} + \lambda_1\Omega_r(\Phi_t^\top \Phi_t+ \lambda_1\mathbf{I})^{-1}\Omega_r^\top\right]^{-1} \\
            &= \lambda_1^{-1} \left[\Omega_r(\Phi_t^\top \Phi_t+ \lambda_1\mathbf{I})^{-1}\Omega_r^\top + \frac{\lambda_2} {\lambda_1} \mathbf{I}\right]^{-1} \\
            V_1 & = \Omega_r^\top - \Phi_t^\top \mathbf{P}^{-1} \mathbf{Q}\\
            & = \Omega_r^\top - \Phi_t^\top (\Phi_t\Phi_t^\top + \lambda_1\mathbf{I})^{-1} \Phi_t \Omega_r^\top \\
            &=\left[ \mathbf{I} - \Phi_t^\top (\Phi_t\Phi_t^\top + \lambda_1\mathbf{I})^{-1} \Phi_t\right] \Omega_r^\top \\
            & = \left[ \mathbf{I} - (\Phi_t^\top\Phi_t + \lambda_1\mathbf{I})^{-1} \Phi_t^\top\Phi_t\right] \Omega_r^\top \\
            & = \left[ \mathbf{I} - (\Phi_t^\top\Phi_t + \lambda_1\mathbf{I})^{-1} (\Phi_t^\top\Phi_t +\lambda_1 \mathbf{I} - \lambda_1 \mathbf{I})\right] \Omega_r^\top \\
            & = \lambda_1(\Phi_t^\top\Phi_t + \lambda_1\mathbf{I})^{-1} \Omega_r^\top \\
            V_2 & = \widetilde{\mathbf{C}}\Phi_t  + \widetilde{\mathbf{D}}\Omega_r \\
            & = -\mathbf{M}\mathbf{R}\mathbf{P}^{-1} + \mathbf{M}\Omega_r \\
            & = \mathbf{M}(\Omega_r - \mathbf{R}\mathbf{P}^{-1}\Phi_t)\\
            & = \mathbf{M}\left[\Omega_r - \Omega_r \Phi_t^\top (\Phi_t \Phi_t^\top + \lambda_1\mathbf{I})^{-1}\Phi_t\right] \\
            & = \mathbf{M}\Omega_r \left[ \mathbf{I} - \Phi_t^\top (\Phi_t \Phi_t^\top + \lambda_1\mathbf{I})^{-1}\Phi_t\right]\\
            & = \lambda_1 \mathbf{M} \Omega_r (\Phi_t^\top \Phi_t + \lambda_1\mathbf{I})^{-1} \\ 
            & = \left[\Omega_r(\Phi_t^\top \Phi_t+ \lambda_1\mathbf{I})^{-1}\Omega_r^\top + \frac{\lambda_2} {\lambda_1} \mathbf{I}\right]^{-1} \Omega_r (\Phi_t^\top \Phi_t + \lambda_1\mathbf{I})^{-1}
    \end{align}
    Then we have, 
    \begin{align}
            \mathbf{V} &= \boldsymbol{\boldsymbol{\xi}_t}^\top \mathbf{\widehat{K}}_\mathrm{PINN}^{-1} \boldsymbol{\boldsymbol{\xi}_t}  \\
            & = (\Phi_t^\top\Phi_t+\lambda_1 \mathbf{I})^{-1} \Phi_t^\top\Phi_t + V_1V_2\\
            & = (\Phi_t^\top\Phi_t+\lambda_1 \mathbf{I})^{-1} \Phi_t^\top\Phi_t + \lambda_1(\Phi_t^\top\Phi_t + \lambda_1\mathbf{I})^{-1} \Omega_r^\top \left[\Omega_r(\Phi_t^\top \Phi_t+ \lambda_1\mathbf{I})^{-1}\Omega_r^\top + \frac{\lambda_2} {\lambda_1} \mathbf{I}\right]^{-1} \Omega_r (\Phi_t^\top \Phi_t + \lambda_1\mathbf{I})^{-1} \\
            &= \mathbf{I} - \lambda_1(\Phi_t^\top\Phi_t+\lambda_1 \mathbf{I})^{-1} + \lambda_1(\Phi_t^\top\Phi_t + \lambda_1\mathbf{I})^{-1} \Omega_r^\top \left[\Omega_r(\Phi_t^\top \Phi_t+ \lambda_1\mathbf{I})^{-1}\Omega_r^\top + \frac{\lambda_2} {\lambda_1} \mathbf{I}\right]^{-1} \Omega_r (\Phi_t^\top \Phi_t + \lambda_1\mathbf{I})^{-1} \\
            &  = \mathbf{I} -  \lambda_1(\Phi_t^\top\Phi_t+\lambda_1 \mathbf{I})^{-1}\left[\mathbf{I} - \Omega_r^\top \left(\Omega_r(\Phi_t^\top \Phi_t+ \lambda_1\mathbf{I})^{-1}\Omega_r^\top+\frac{\lambda_2} {\lambda_1} \mathbf{I}\right)^{-1} \Omega_r (\Phi_t^\top\Phi_t+\lambda_1 \mathbf{I})^{-1}\right] \\ 
            & = \mathbf{I} -  \lambda_1(\Phi_t^\top\Phi_t+\lambda_1 \mathbf{I})^{-1}\left[\mathbf{I} -  \left(\Omega_r^\top\Omega_r(\Phi_t^\top \Phi_t+ \lambda_1\mathbf{I})^{-1}+\frac{\lambda_2} {\lambda_1} \mathbf{I}\right)^{-1} \Omega_r^\top\Omega_r (\Phi_t^\top\Phi_t+\lambda_1 \mathbf{I})^{-1}\right] \\ 
            & = \mathbf{I} -  \lambda_1(\Phi_t^\top\Phi_t+\lambda_1 \mathbf{I})^{-1}\left[\mathbf{I} -  \mathbf{I} + \frac{\lambda_2}{\lambda_1}\left(\Omega_r^\top\Omega_r(\Phi_t^\top \Phi_t+ \lambda_1\mathbf{I})^{-1}+\frac{\lambda_2} {\lambda_1} \mathbf{I}\right)^{-1} \right] \\ 
            & = \mathbf{I} - \lambda_2 (\Phi_t^\top \Phi_t+ \lambda_1\mathbf{I})^{-1}\left(\Omega_r^\top\Omega_r(\Phi_t^\top \Phi_t+ \lambda_1\mathbf{I})^{-1}+\frac{\lambda_2} {\lambda_1} \mathbf{I}\right)^{-1} \\
            & =\mathbf{I} - \lambda_2 \left(\Omega_r^\top\Omega_r+\frac{\lambda_2} {\lambda_1} (\Phi_t^\top \Phi_t+ \lambda_1\mathbf{I})\right)^{-1} \\
            & =\mathbf{I} - \lambda_2 \left(\Omega_r^\top\Omega_r+\frac{\lambda_2} {\lambda_1} \Phi_t^\top \Phi_t+ \lambda_2 \mathbf{I}\right)^{-1}  
    \end{align}
In conclusion, we have 
\begin{align}
\label{Eqn:xi_t-K-xi}
        \boldsymbol{\xi}_t^\top \mathbf{\widehat{K}}_\mathrm{PINN}^{-1} \boldsymbol{\xi}_t &= \mathbf{I} - \lambda_2 \left(\Omega_r^\top\Omega_r+\frac{\lambda_2} {\lambda_1} \Phi_t^\top \Phi_t+ \lambda_2 \mathbf{I}\right)^{-1} 
        \\
        &= \mathbf{I} -  \left(\frac{\Omega_r^\top\Omega_r}{\lambda_2}+\frac{\Phi_t^\top \Phi_t} {\lambda_1} + \mathbf{I}\right)^{-1} 
\end{align}

Replace Eqn. \ref{Eqn:xi_t-K-xi} to the expression of $\textup{Cov}(f_t \rvert \mathbf{Y}_t; \mathbf{U}_r)$ in Eqn. \ref{Eqn:covar_f_condition_on_y_and_u}, we have 

\begin{equation}
\label{Eqn:covar_f_condition_on_y_and_u_final}
    \textup{Cov}(f \rvert \mathbf{Y}_t; \mathbf{U}_r) = \nu_t^2\Phi_t \left(\frac{\Omega_r^\top\Omega_r}{\lambda_2}+\frac{\Phi_t^\top \Phi_t} {\lambda_1} + \mathbf{I}\right)^{-1} \Phi_t^\top
\end{equation}
Combining the expression of $\textup{Cov}(f\rvert \mathbf{U}_r)$ in Eqn. \ref{Eqn:covar_f_condition_on_u} with Eqn. \ref{Eqn:covar_f_condition_on_y_and_u_final}, with $H$ is the entropy function, we have
\begin{align}
     I(f;\mathbf{Y}_t \rvert \mathbf{U}_r) &= H(f \rvert \mathbf{U}_r) - H(f \rvert \mathbf{Y}_t; \mathbf{U}_r) \\
     & = \frac{1}{2}\log\det (\textup{Cov}(f \rvert \mathbf{U}_r)) - \frac{1}{2}\log\det (\textup{Cov}(f \rvert \mathbf{Y}_t;  \mathbf{U}_r)) \\
     & = \frac{1}{2}\log \det (\nu_t^2\Phi_t \left(\frac{\Omega_r^\top \Omega_r}{\lambda_2} + \mathbf{I}\right)^{-1} \Phi_t^\top) - \frac{1}{2}\log\det (\nu_t^2\Phi_t \left(\frac{\Omega_r^\top\Omega_r}{\lambda_2}+\frac{\Phi_t^\top \Phi_t} {\lambda_1} + \mathbf{I}\right)^{-1} \Phi_t^\top) \\
\label{Eqn:conditional_information_f_and_Yt_give_Ur}
     &= \frac{1}{2} \log \frac{\det(\Phi_t \left(\frac{\Omega_r^\top \Omega_r}{\lambda_2} + \mathbf{I}\right)^{-1} \Phi_t^\top)}{\det(\Phi_t \left(\frac{\Omega_r^\top\Omega_r}{\lambda_2}+\frac{\Phi_t^\top \Phi_t} {\lambda_1} + \mathbf{I}\right)^{-1} \Phi_t^\top)}
\end{align}

Then we have 
\begin{align}
    I (f; \mathbf{Y}_t; \mathbf{U}_r) &= I(f; \mathbf{Y}_t) - I (f; \mathbf{Y}_t \rvert \mathbf{U}_r) \\
        &= \frac{1}{2} \log \det (\frac{\Phi_t\Phi_t^\top}{\lambda_1} +\mathbf{I}) - \frac{1}{2} \log \frac{\det(\Phi_t \left(\frac{\Omega_r^\top \Omega_r}{\lambda_2} + \mathbf{I}\right)^{-1} \Phi_t^\top)}{\det(\Phi_t \left(\frac{\Omega_r^\top\Omega_r}{\lambda_2}+\frac{\Phi_t^\top \Phi_t} {\lambda_1} + \mathbf{I}\right)^{-1} \Phi_t^\top)} 
        \\
        & = \frac{1}{2}\log \frac{ \det (\frac{\Phi_t\Phi_t^\top}{\lambda_1} +\mathbf{I})
        \det(\Phi_t \left(\frac{\Omega_r^\top\Omega_r}{\lambda_2}+\frac{\Phi_t^\top \Phi_t} {\lambda_1} + \mathbf{I}\right)^{-1} \Phi_t^\top)
        }{\det(\Phi_t \left(\frac{\Omega_r^\top \Omega_r}{\lambda_2} + \mathbf{I}\right)^{-1} \Phi_t^\top)} 
        \\ 
        & = \frac{1}{2}\log \frac{\det[\left(\frac{\Phi_t\Phi_t^\top}{\lambda_1} +\mathbf{I}\right) \Phi_t \left(\frac{\Omega_r^\top\Omega_r}{\lambda_2}+\frac{\Phi_t^\top \Phi_t} {\lambda_1} + \mathbf{I}\right)^{-1}\Phi_t^\top]}
        {\det[ \Phi_t \left(\frac{\Omega_r^\top \Omega_r}{\lambda_2} + \mathbf{I}\right)^{-1} \Phi_t^\top]} 
        \\ 
        & = \frac{1}{2}\log \frac{
        \det[\Phi_t \left(\frac{\Phi_t^\top\Phi_t}{\lambda_1} +\mathbf{I}\right)  \left(\frac{\Omega_r^\top\Omega_r}{\lambda_2}+\frac{\Phi_t^\top \Phi_t} {\lambda_1} + \mathbf{I}\right)^{-1}\Phi_t^\top]}
        {\det[ \Phi_t \left(\frac{\Omega_r^\top \Omega_r}{\lambda_2} + \mathbf{I}\right)^{-1} \Phi_t^\top]} 
        \\
        & = \frac{1}{2}\log \frac{
        \det[\Phi_t  \left(\frac{\Omega_r^\top \Omega_r}{\lambda_2} \left(\frac{\Phi_t^\top\Phi_t}{\lambda_1} +\mathbf{I}\right)^{-1} + \mathbf{I}\right)^{-1}\Phi_t^\top]}
        {\det[ \Phi_t \left(\frac{\Omega_r^\top \Omega_r}{\lambda_2} + \mathbf{I}\right)^{-1} \Phi_t^\top]}
        \\
        \label{Eqn:det_division}
        &= \frac{1}{2}\log \frac{\det\left(\frac{\Omega_r^\top \Omega_r}{\lambda_2} \left(\frac{\Phi_t^\top\Phi_t}{\lambda_1} +\mathbf{I}\right)^{-1} + \mathbf{I}\right)^{-1}}{\det (\frac{\Omega_r^\top\Omega_r}{\lambda_2} + \mathbf{I})^{-1}}
        \\
        & = \frac{1}{2}\log \frac{\det (\frac{\Omega_r^\top\Omega_r}{\lambda_2} + \mathbf{I})}{\det\left(\frac{\Omega_r^\top \Omega_r}{\lambda_2} \left(\frac{\Phi_t^\top\Phi_t}{\lambda_1} +\mathbf{I}\right)^{-1} + \mathbf{I}\right)}
        \\ 
        & = \frac{1}{2}  \log \left(\frac{\det(\frac{\Phi_t^\top \Phi_t}{\lambda_1} + \mathbf{I})\det(\frac{\Omega_r^\top \Omega_r}{\lambda_2} + \mathbf{I})}{\det(\frac{\Phi_t^\top \Phi_t}{\lambda_1} + \frac{\Omega_r^\top \Omega_r}{\lambda_2} + \mathbf{I})} \right), 
\end{align}
where Eqn \ref{Eqn:det_division} is resulted from technical Lemma \ref{lemma:det_division}. 
\end{proof}

\subsection{Proof of Lemma \ref{lemma:confidence_bound}}
\ConfidenceBound*
\textbf{To prove Lemma \ref{lemma:confidence_bound}, we utilize the following lemmas:}

\begin{sublemma}[Theorem 1 \cite{zhou2008derivative})]
\label{lemma:rkhs_derivative}
    Let $s \in \mathbb{N}$ and $k \colon X \times X \rightarrow \mathbb{R}$ be a Mercer kernel such that $k \in C^{2s} (X \times X)$. Denote $D^\alpha f(\mathbf{x}), \alpha \in \mathbb{Z}^n_+$ is the partial derivative of $f$ at $\mathbf{x} \in \mathbb{R}^n$ (if it exists) as:
\begin{align}
    D^\alpha f(\mathbf{x}) =  \frac{\partial ^ {\lvert \alpha \rvert}}{\partial{\mathbf{x}_1^{\alpha_1}} \dots \partial{\mathbf{x}_n^{\alpha_n}}} f(\mathbf{x}),
\end{align}
where $\lvert \alpha \rvert = \sum_{j=1}^n \alpha_j$. Denote $\mathbb{I}_s = \{\alpha \in \mathbb{Z}_+^n, \lvert \alpha \rvert \le s\}$ and  $(D^\alpha k)_\mathbf{x}$ as the function on X given by $(D^\alpha k)_\mathbf{x} (\mathbf{y}) = D^\alpha k(\mathbf{x}, \mathbf{y}) =  \frac{\partial^ {\lvert \alpha \rvert}}{\partial{\mathbf{x}_1^{\alpha_1}} \dots \partial{\mathbf{x}_n^{\alpha_n}}} k(\mathbf{x}, \mathbf{y})$.

Then the following statements hold: 

\begin{enumerate}
    \item For any $\mathbf{x} \in X$ and  $\alpha \in \mathbb{I}_s$, $(D^\alpha k)_\mathbf{x} \in \mathcal{H}_k$.
    \item A partial derivative reproducing property holds true for $\alpha \in \mathbb{I}_s$
    \begin{align}
        D^\alpha f(\mathbf{x}) = \langle (D^\alpha k)_\mathbf{x}, f \rangle,  \forall \mathbf{x} \in X, f \in \mathcal{H}_k
    \end{align}
\end{enumerate}
\end{sublemma}

\begin{sublemma}[Theorem 1 \cite{abbasi2011improved}]
\label{lemma:self-normalized_bound}
Let $\{\mathcal{F}_t\}_{t=0}^\infty$ be a filtration. Let $\{\eta_t\}_{t=0}^\infty$ be a real-valued stochastic process such that $\eta_t$ is $\mathcal{F}_t$-measurable and $\eta_t$ is conditionally
$R$-sub-Gaussian for some $R \ge 0$. Let $\{X_t\}_{t=0}^\infty$ be an $\mathbb{R}^d$-valued stochastic process such that $X_t$ is $\mathcal{F}_{t-1}$-measurable. Assume that $\mathbf{V}$
is a d × d positive definite matrix. For any $t \ge 0$, define
\begin{align}
    \overline{\mathbf{V}}_t &= \mathbf{V} + \sum_{s=1}^t X_s X_s^\top  \\
    S_t &= \sum_{s=1}^t \eta_s X_s
\end{align}
Then, for any $\delta > 0$, with probability at least $1 - \delta$, for all $t \ge 0$,
\[
\norm{S_t}^2_{\overline{\mathbf{V}}_t^{-1}} \le 2R^2\log(\frac{\det(\overline{\mathbf{V}}_t)^{1/2}\det(\mathbf{V})^{-1/2}}{\delta})
\] 
\end{sublemma}
\begin{sublemma}
\label{lemma:IG_with_pde_upper_bound}
    Assume that $\norm{\omega(\cdot)}_2 \le L$, where $\omega(\cdot)  = \nabla_{\boldsymbol{\theta}} \mathcal{N}[h] (\cdot, \boldsymbol{\theta}_0)$ defined in Section 5 of the main paper, and denote $\rho_{min}(\mathbf{K}_{uu})$ is the smallest eigenvalue of kernel matrix $\mathbf{K}_{uu}$ defined in lemma \ref{lemma:PINN_GP}. Then we have:
\begin{align}
    \frac{1}{2}\log \det(\frac{\Phi_t^\top \Phi_t}{\lambda_1} + \frac{\Omega_r^\top \Omega_r}{\lambda_2} + \mathbf{I}) &\le  \frac{1}{2}\log\det (\frac{\mathbf{K}_{uu}}{\lambda_1} +\mathbf{I})+\frac{N_r L^2}{2(1+ \rho_{min}(\mathbf{K}_{uu})/\lambda_1)}\\
    & \le \gamma_t +  \frac{N_r  L^2}{2(1+ \rho_{min}(\mathbf{K}_{uu})/\lambda_1)}
\end{align}
\end{sublemma}
\begin{proof}[Proof of Lemma \ref{lemma:IG_with_pde_upper_bound}]
\begin{align}
            \frac{1}{2}\log \det(\frac{\Phi_t^\top \Phi_t}{\lambda_1} + \frac{\Omega_r^\top \Omega_r}{\lambda_2} + \mathbf{I})
            & = \frac{1}{2}\log \det(\frac{\Phi_t^\top \Phi_t}{\lambda_1} +  \mathbf{I} + \frac{1}{\lambda_2} \sum_{j=1}^{N_r} \omega(\mathbf{z}_j) \omega(\mathbf{z}_j)^\top) 
            \\
            & = \frac{1}{2}\log \det( \underbrace{\frac{\Phi_t^\top \Phi_t}{\lambda_1} +  \mathbf{I} + \frac{1}{\lambda_2} \sum_{j=1}^{N_r-1} \omega(\mathbf{z}_j) \omega(\mathbf{z}_j)^\top}_{\mathbf{K}_{r-1}} + \frac{1}{\lambda_2} \omega(\mathbf{z}_{N_r}) \omega(\mathbf{z}_{N_r})^\top) 
            \\
             & = \frac{1}{2}\log \left[ \det(\mathbf{K}_{r-1}) \det(\mathbf{I} + \mathbf{K}_{r-1}^{-1/2}\omega(\mathbf{z}_{N_r}) \omega(\mathbf{z}_{N_r})^\top(\mathbf{K}_{r-1}^{-1/2})^\top) \right]
             \\
             & = \frac{1}{2}\log \left[ \det(\mathbf{K}_{r-1}) (1+\norm{\omega(\mathbf{z}_{N_r})}^2_{\mathbf{K}^{-1}_{r-1}}) \right] 
             \\
             & = \frac{1}{2}\log \left[ \det(\frac{\Phi_t^\top \Phi_t}{\lambda_1} + \mathbf{I}) \prod_{j=1}^{N_r}(1+\norm{\omega(\mathbf{z}_j)}^2_{\mathbf{K}^{-1}_{j-1}}) \right] 
             \\
             & = \frac{1}{2}\log \det(\frac{\Phi_t^\top \Phi_t}{\lambda_1} + \mathbf{I}) + \frac{1}{2} \sum_{j=1}^{N_r} \log(1+\norm{\omega(\mathbf{z}_j)}^2_{\mathbf{K}^{-1}_{j-1}}) 
             \\
             & = \frac{1}{2}\log \det(\frac{\mathbf{K}_{uu}}{\lambda_1} + \mathbf{I}) + \frac{1}{2} \sum_{j=1}^{N_r} \log(1+\norm{\omega(\mathbf{z}_j)}^2_{\mathbf{K}^{-1}_{j-1}})
             \\
             & \le \gamma_t + \frac{1}{2} \sum_{j=1}^{N_r} \norm{\omega(\mathbf{z}_j)}^2_{\mathbf{K}^{-1}_{j-1}} 
             \\
             &  \le \gamma_t + \frac{1}{2} \sum_{j=1}^{N_r} \rho_{max}(\mathbf{K}^{-1}_{j-1}) \norm{\omega(\mathbf{z}_j)}_2^2
             \\
             & \le \gamma_t + \frac{1}{2}\sum_{j=1}^{N_r} \rho_{min}^{-1}(\mathbf{K}_{j-1}) \norm{\omega(\mathbf{z}_j)}_2^2
             \\
             & \le \gamma_t + \frac{N_r L^2}{2 \rho_{min}\left(\frac{\mathbf{K}_{uu}}{\lambda_1} + \mathbf{I}\right)} 
             \\
             &= \gamma_t + \frac{N_r L^2}{2(1+ \rho_{min}(\mathbf{K}_{uu})/\lambda_1)}
    \end{align}
\end{proof}
The fourth equality leveraged the property that a matrix in the form of $\mathbf{I} + vv^\top$ has an eigenvalue of $1 + \lVert v \rVert^2$, associated with the eigenvector $v$, while all other eigenvalues are equal to 1. The 7th equality utilized Weinstein–Aronszajn identity, which indicates that $\det(\frac{\Phi_t^\top \Phi_t}{\lambda_1} + \mathbf{I}) = \det(\frac{\mathbf{K}_{uu}}{\lambda_1} + \mathbf{I})$. The first inequality is from $\log(1+a) \le a$ and $\gamma_t := \max_{\mathcal{A} \subset \mathcal{D}: \lvert \mathcal{A} \rvert=t} I(\mathbf{Y}_\mathcal{A}; f_\mathcal{A}) \ge \frac{1}{2}\log \det(\frac{\mathbf{K}_{uu}}{\lambda_1} + \mathbf{I})$, while the second inequality uses the min-max theorem for Hermitian matrices. The last equality used properties of eigenvalues of shifted matrices. 

\paragraph{Extra Definitions}
Let us define terms for the convenience as follows:
\begin{align}
     \mathbf{F}_t & = (f(\mathbf{x}_1), \cdots, f(\mathbf{x}_t))^\top 
     \\
      \mathbf{G}_r & = (g(\mathbf{z}_1), \cdots, g(\mathbf{z}_{N_r}))^\top \\
     \boldsymbol{\epsilon}_t & = (y_1 - f(\mathbf{x}_1), \cdots, y_t - f(\mathbf{x}_t))^\top \\
     \boldsymbol{\eta}_r & = (u_1 - g(\mathbf{z}_1), \cdots, u_{N_r} - g(\mathbf{z}_{N_r}))^\top,
\end{align}
where $\{y_i\}_{i=1}^T$ and $\{u_j\}_{j=1}^{N_r}$ are defined in Lemma \ref{lemma:PINN_mean_cov}. Now we are ready to prove Lemma \ref{lemma:confidence_bound}.

\begin{proof}[Proof of Lemma \ref{lemma:confidence_bound}]
From Lemma \ref{lemma:PINN_mean_cov}, we have $\mu_t^f(\mathbf{x}) =\phi(\mathbf{x})^\top \boldsymbol{\xi}_t^\top \mathbf{\widehat{K}}_\mathrm{PINN}^{-1} \begin{bmatrix}
         \mathbf{Y}_t \\
         \mathbf{U}_r\end{bmatrix}$. Therefore, the confidence interval is then expressed as:
\begin{align}
        \left\lvert f(\mathbf{x}) - \mu_t^f(\mathbf{x}) \right\rvert &= \left\lvert f(\mathbf{x}) - \phi(\mathbf{x})^\top \boldsymbol{\xi}_t^\top \mathbf{\widehat{K}}_\mathrm{PINN}^{-1} \begin{bmatrix}
         \mathbf{Y}_t \\
         \mathbf{U}_r\end{bmatrix} \right \rvert
        \\
        & = \left\lvert f(\mathbf{x}) - \phi(\mathbf{x})^\top \begin{bmatrix}
            \Phi_t^\top \widetilde{\mathbf{A}}\mathbf{Y}_t + \Omega_r^\top \widetilde{\mathbf{C}}\mathbf{Y}_t + \Phi_t^\top \widetilde{\mathbf{B}}\mathbf{U}_r + \Omega_r^\top \widetilde{\mathbf{D}}\mathbf{U}_r
        \end{bmatrix} \right\rvert \\
        & = \left\lvert f(\mathbf{x}) - \phi(\mathbf{x})^\top \begin{bmatrix}
            \Phi_t^\top \widetilde{\mathbf{A}}(\mathbf{F}_t + \boldsymbol{\epsilon}_t) + \Omega_r^\top \widetilde{\mathbf{C}}(\mathbf{F}_t + \boldsymbol{\epsilon}_t) + \Phi_t^\top \widetilde{\mathbf{B}}(\mathbf{G}_r +\boldsymbol{\eta}_r) + \Omega_r^\top \widetilde{\mathbf{D}}(\mathbf{G}_r +\boldsymbol{\eta}_r)
        \end{bmatrix} \right\rvert \\
        & \le \underbrace{\left\lvert f(\mathbf{x}) - \phi(\mathbf{x})^\top \begin{bmatrix}
            \Phi_t^\top \widetilde{\mathbf{A}}\mathbf{F}_t + \Omega_r^\top \widetilde{\mathbf{C}}\mathbf{F}_t + \Phi_t^\top \widetilde{\mathbf{B}}\mathbf{G}_r+ \Omega_r^\top \widetilde{\mathbf{D}}\mathbf{G}_r
        \end{bmatrix} \right\rvert}_{T_1} \\ 
        & + \underbrace{\left\lvert \phi(\mathbf{x})^\top \Phi_t^\top \widetilde{\mathbf{A}}\boldsymbol{\epsilon}_t +  \phi(\mathbf{x})^\top \Omega_r^\top \widetilde{\mathbf{C}}\boldsymbol{\epsilon}_t +  \phi(\mathbf{x})^\top \Phi_t^\top \widetilde{\mathbf{B}}\boldsymbol{\eta}_r) + \phi(\mathbf{x})^\top \Omega_r^\top \widetilde{\mathbf{D}}\boldsymbol{\eta}_r)\right\rvert}_{T_2}
    \end{align}
    
    \paragraph{Bound term $T_1 = \left\lvert f(\mathbf{x}) - \phi(\mathbf{x})^\top \begin{bmatrix}
            \Phi_t^\top \widetilde{\mathbf{A}}\mathbf{F}_t + \Omega_r^\top \widetilde{\mathbf{C}}\mathbf{F}_t + \Phi_t^\top \widetilde{\mathbf{B}}\mathbf{G}_r + \Omega_r^\top \widetilde{\mathbf{D}}\mathbf{G}_r
        \end{bmatrix} \right\rvert$.\\}
    
    As the unknown reward function $f$ lies in the RKHS with NTK kernel and $\phi(\cdot)$ can be considered as the finite approximation of the linear feature map w.r.t NTK kernel. Then we have $f(\mathbf{x}) = \langle f, \phi(\mathbf{x}) \rangle$. Next, it is clear that Lemma \ref{lemma:rkhs_derivative} can be extended to any linear differential operator. Then we have $g(\mathbf{x}) = \langle g, \omega(\mathbf{x}) \rangle$, where $\omega(\mathbf{x}) = \nabla_{\boldsymbol{\theta}}\mathcal{N}[h] (\mathbf{x}, \boldsymbol{\theta}_0)$ as defined in the main paper. Therefore, 
    \begin{align}
            T_1 &= \left\lvert f(\mathbf{x}) - \phi(\mathbf{x})^\top \begin{bmatrix}
            \Phi_t^\top \widetilde{\mathbf{A}}\mathbf{F}_t + \Omega_r^\top \widetilde{\mathbf{C}}\mathbf{F}_t + \Phi_t^\top \widetilde{\mathbf{B}}\mathbf{G}_r + \Omega_r^\top \widetilde{\mathbf{D}}\mathbf{G}_r
        \end{bmatrix} \right\rvert \\
                &=  \left\lvert \phi(\mathbf{x})^\top f - \phi(\mathbf{x})^\top \begin{bmatrix}
            \Phi_t^\top \widetilde{\mathbf{A}}\Phi_tf + \Omega_r^\top \widetilde{\mathbf{C}}\Phi_tf + \Phi_t^\top \widetilde{\mathbf{B}}\Omega_rf + \Omega_r^\top \widetilde{\mathbf{D}}\Omega_rf
        \end{bmatrix} \right\rvert \\
                &= \left\lvert \phi(\mathbf{x})^\top f - \phi(\mathbf{x})^\top \begin{bmatrix}
            \Phi_t^\top \widetilde{\mathbf{A}}\Phi_t + \Omega_r^\top \widetilde{\mathbf{C}}\Phi_t + \Phi_t^\top \widetilde{\mathbf{B}}\Omega_r + \Omega_r^\top \widetilde{\mathbf{D}}\Omega_r
        \end{bmatrix} f \right\rvert \\
                & = \left\lvert \phi(\mathbf{x})^\top f - \phi(\mathbf{x})^\top 
            \boldsymbol{\boldsymbol{\xi}_t}^\top \mathbf{\widehat{K}}_\mathrm{PINN}^{-1} \boldsymbol{\boldsymbol{\xi}_t} f \right\rvert  \\
            & \le \norm{ \phi(\mathbf{x})^\top - \phi(\mathbf{x})^\top 
            \boldsymbol{\boldsymbol{\xi}_t}^\top \mathbf{\widehat{K}}_\mathrm{PINN}^{-1} \boldsymbol{\boldsymbol{\xi}_t}}_{k_\textup{NTK-PINN}}\norm{f}_{k_\textup{NTK-PINN}}\\
            & = \norm{f}_{k_\textup{NTK-PINN}} \norm{\phi(\mathbf{x})^\top(\mathbf{I} - \mathbf{V}) }_{k_\textup{NTK-PINN}} \\
            & = \norm{f}_{k_\textup{NTK-PINN}} \norm{\lambda_2\phi(\mathbf{x})^\top \left(\underbrace{\Omega_r^\top\Omega_r+\frac{\lambda_2} {\lambda_1} \Phi_t^\top \Phi_t}_{\mathbf{Z}} + \lambda_2 \mathbf{I}\right)^{-1} }_{k_\textup{NTK-PINN}} \\
            & \le \norm{f}_{k_\textup{NTK-PINN}} \sqrt{\lambda_2\phi(\mathbf{x})^\top (\mathbf{Z} + \lambda_2\mathbf{I})^{-1} \lambda_2 \mathbf{I} (\mathbf{Z} + \lambda_2\mathbf{I})^{-1} \phi(\mathbf{x})} \\
            & \le \norm{f}_{k_\textup{NTK-PINN}} \sqrt{\lambda_2 \phi(\mathbf{x})^\top (\mathbf{Z} + \lambda_2\mathbf{I})^{-1}  (\lambda_2 \mathbf{I} + \mathbf{Z})(\mathbf{Z} + \lambda_2\mathbf{I})^{-1} \phi(\mathbf{x})}\\
            & = \norm{f}_{k_\textup{NTK-PINN}} \sqrt{\lambda_2 \phi(\mathbf{x})^\top (\mathbf{Z} + \lambda_2\mathbf{I})^{-1} \phi(\mathbf{x})} \\
            & = B \sigma_t^f(\mathbf{x})
    \end{align}
    The first inequality is by Cauchy-Schwartz while the third inequality uses the fact that matrix $\mathbf{Z} = \Omega_r^\top\Omega_r+\frac{\lambda_2} {\lambda_1} \Phi_t^\top \Phi_t$ is positive semi-definite. The fifth equality inherits the expression of term $\mathbf{V} = \boldsymbol{\xi}_t ^\top \mathbf{\widehat{K}}_\textup{PINN}^{-1} \boldsymbol{\xi}_t$ calculated in line (\ref{Eqn:xi_t-K-xi}). The last equality is from the fact that:
    \begin{align}
    \label{Eqn:sigma_f_by_V}
    \left(\sigma^f_t\right)^2(\mathbf{x}) & = \langle \phi(\mathbf{x}),  \phi(\mathbf{x}) \rangle - \phi(\mathbf{x})^\top \boldsymbol{\boldsymbol{\xi}_t}^\top \mathbf{\widehat{K}}_\mathrm{PINN}^{-1} \boldsymbol{\boldsymbol{\xi}_t} \phi(\mathbf{x}) \\ 
            & = \phi(\mathbf{x})^\top \left[\mathbf{I} - \phi(\mathbf{x})^\top \boldsymbol{\boldsymbol{\xi}_t}^\top \mathbf{\widehat{K}}_\mathrm{PINN}^{-1} \boldsymbol{\boldsymbol{\xi}_t}\right] \phi(\mathbf{x})\\
            & = \phi(\mathbf{x})^\top (\mathbf{I} - \mathbf{V}) \phi(\mathbf{x}) \\
            &  = \lambda_2 \phi(\mathbf{x})^\top\left(\Omega_r^\top\Omega_r+\frac{\lambda_2} {\lambda_1} \Phi_t^\top \Phi_t+ \lambda_2 \mathbf{I}\right)^{-1} \phi(\mathbf{x}) 
    \end{align}
    \paragraph{Bound term $T_2=  \left\lvert \phi(\mathbf{x})^\top \Phi_t^\top \widetilde{\mathbf{A}}\boldsymbol{\epsilon}_t +  \phi(\mathbf{x})^\top \Omega_r^\top \widetilde{\mathbf{C}}\boldsymbol{\epsilon}_t +  \phi(\mathbf{x})^\top \Phi_t^\top \widetilde{\mathbf{B}}\boldsymbol{\eta}_r) + \phi(\mathbf{x})^\top \Omega_r^\top \widetilde{\mathbf{D}}\boldsymbol{\eta}_r) \right\rvert$.\\} 
    
    We start to calculate term $T_2$ as follows:
    \begin{align}
            T_2 &=\left\lvert \phi(\mathbf{x})^\top \Phi_t^\top \widetilde{\mathbf{A}}\boldsymbol{\epsilon}_t +  \phi(\mathbf{x})^\top \Omega_r^\top \widetilde{\mathbf{C}}\boldsymbol{\epsilon}_t +  \phi(\mathbf{x})^\top \Phi_t^\top \widetilde{\mathbf{B}}\boldsymbol{\eta}_r) + \phi(\mathbf{x})^\top \Omega_r^\top \widetilde{\mathbf{D}}\boldsymbol{\eta}_r) \right\rvert \\
            &= \phi(\mathbf{x})^\top\left[\underbrace{(\Phi_t^\top \widetilde{\mathbf{A}} +  \Omega_r^\top \widetilde{\mathbf{C}})\boldsymbol{\epsilon}_t + (\Phi_t^\top\widetilde{\mathbf{B}} + \Omega_r^\top\widetilde{\mathbf{D}}) \boldsymbol{\eta}_r)}_{T_{2a}} \right] \\
    \end{align}
   We need to calculate $T_{2a}$:
   \begin{align}
       T_{2a} &= (\Phi_t^\top \widetilde{\mathbf{A}} +  \Omega_r^\top \widetilde{\mathbf{C}})\boldsymbol{\epsilon}_t + (\Phi_t^\top\widetilde{\mathbf{B}} + \Omega_r^\top\widetilde{\mathbf{D}}) \boldsymbol{\eta}_r) \\
       & = \left[\Phi_t^\top (\mathbf{P}^{-1} - \mathbf{P}^{-1}\mathbf{Q}\widetilde{\mathbf{C}}) +  \Omega_r^\top \widetilde{\mathbf{C}}\right]\boldsymbol{\epsilon}_t + (\Omega_r^\top - \Phi_t^\top\mathbf{P}^{-1}\mathbf{Q})\widetilde{\mathbf{D}} \boldsymbol{\eta}_r) \\ 
       & = \Phi_t^\top \mathbf{P}^{-1} \Phi_t \boldsymbol{\epsilon}_t + \underbrace{(\Omega_r^\top - \Phi_t^\top\mathbf{P}^{-1}\mathbf{Q})}_{V_1}(\widetilde{\mathbf{C}} \boldsymbol{\epsilon}_t + \widetilde{\mathbf{D}} \boldsymbol{\eta}_r) ) \\
       & = \Phi_t^\top (\Phi_t\Phi_t^\top + \lambda_1\mathbf{I})^{-1} \boldsymbol{\epsilon}_t + \lambda_1(\Phi_t^\top\Phi_t + \lambda_1\mathbf{I})^{-1} \Omega_r^\top (\widetilde{\mathbf{C}} \boldsymbol{\epsilon}_t + \widetilde{\mathbf{D}} \boldsymbol{\eta}_r) ) \\
       & = (\Phi_t^\top\Phi_t + \lambda_1\mathbf{I})^{-1} \Phi_t^\top \boldsymbol{\epsilon}_t + \lambda_1(\Phi_t^\top\Phi_t + \lambda_1\mathbf{I})^{-1} \Omega_r^\top (\widetilde{\mathbf{C}} \boldsymbol{\epsilon}_t + \widetilde{\mathbf{D}} \boldsymbol{\eta}_r) ) \\
       & = (\Phi_t^\top\Phi_t + \lambda_1\mathbf{I})^{-1} (\Phi_t^\top \boldsymbol{\epsilon}_t + \lambda_1\Omega_r^\top \widetilde{\mathbf{C}} \boldsymbol{\epsilon}_t +  \lambda_1\Omega_r\widetilde{\mathbf{D}} \boldsymbol{\eta}_r) \\
       & = (\underbrace{\Phi_t^\top\Phi_t + \lambda_1\mathbf{I}}_\mathbf{U})^{-1} (\underbrace{\Phi_t^\top \boldsymbol{\epsilon}_t + \lambda_1\Omega_r^\top \mathbf{MR}\mathbf{P}^{-1} \boldsymbol{\epsilon}_t -  \lambda_1\Omega_r \mathbf{M} \boldsymbol{\eta}_r)}_{T_{2a}^\prime})\\
   \end{align}
   Then, let $\mathbf{U} = \Phi_t^\top\Phi_t + \lambda_1\mathbf{I}$ and $\mathbf{J} = \Omega_r^\top \Omega_r$, we have 
   \begin{align}
       T_{2a}^\prime &= \Phi_t^\top \boldsymbol{\epsilon}_t -
       \lambda_1\Omega_r^\top \mathbf{MR}\mathbf{P}^{-1} \boldsymbol{\epsilon}_t +  \lambda_1\Omega_r \mathbf{M} \boldsymbol{\eta}_r) \\
       &=\Phi_t^\top \boldsymbol{\epsilon}_t - \Omega_r^\top \left(\Omega_r\mathbf{U}^{-1} \Omega_r^\top + \frac{\lambda_2}{\lambda_1}\mathbf{I}\right)^{-1} \Omega_r \Phi_t^\top (\Phi_t\Phi_t^\top + \lambda_1\mathbf{I})^{-1}\boldsymbol{\epsilon}_t + \Omega_r^\top\left[\Omega_r\mathbf{U}^{-1} \Omega_r^\top + \frac{\lambda_2}{\lambda_1}\mathbf{I}\right]^{-1}\boldsymbol{\eta}_r) \\
       & = \Phi_t^\top \boldsymbol{\epsilon}_t - \Omega_r^\top \Omega_r \left(\mathbf{U}^{-1}\Omega_r^\top \Omega_r + \frac{\lambda_2}{\lambda_1}\mathbf{I} \right)^{-1} \mathbf{U}^{-1} \Phi_t^\top \boldsymbol{\epsilon}_t + \left(\Omega_r^\top \Omega_r\mathbf{U}^{-1} + \frac{\lambda_2}{\lambda_1}\mathbf{I}\right) ^{-1} \Omega_r^\top \boldsymbol{\eta}_r)  \\
       & = \Phi_t^\top \boldsymbol{\epsilon}_t - \mathbf{J} \left(\mathbf{U}^{-1} \mathbf{J} + \frac{\lambda_2}{\lambda_1}\mathbf{I} \right)^{-1} \mathbf{U}^{-1} \Phi_t^\top \boldsymbol{\epsilon}_t + \left(\mathbf{J} \mathbf{U}^{-1} + \frac{\lambda_2}{\lambda_1}\mathbf{I}\right) \Omega_r^\top \boldsymbol{\eta}_r) \\
       & = \Phi_t^\top \boldsymbol{\epsilon}_t -  \left(\mathbf{J}\mathbf{U}^{-1}  + \frac{\lambda_2}{\lambda_1}\mathbf{I} \right)^{-1} \mathbf{J}\mathbf{U}^{-1} \Phi_t^\top \boldsymbol{\epsilon}_t + \left(\mathbf{J} \mathbf{U}^{-1} + \frac{\lambda_2}{\lambda_1}\mathbf{I}\right) \Omega_r^\top \boldsymbol{\eta}_r) \\ 
       & = \left[ \mathbf{I} -\left(\mathbf{J}\mathbf{U}^{-1}  + \frac{\lambda_2}{\lambda_1}\mathbf{I} \right)^{-1} \mathbf{J}\mathbf{U}^{-1} \right] \Phi_t^\top \boldsymbol{\epsilon}_t + \left(\mathbf{J} \mathbf{U}^{-1} + \frac{\lambda_2}{\lambda_1}\mathbf{I}\right) \Omega_r^\top \boldsymbol{\eta}_r) \\
       & = \frac{\lambda_2}{\lambda_1} \left(\mathbf{J}\mathbf{U}^{-1}  + \frac{\lambda_2}{\lambda_1}\mathbf{I} \right)^{-1}\Phi_t^\top \boldsymbol{\epsilon}_t +  \left(\mathbf{J}\mathbf{U}^{-1}  + \frac{\lambda_2}{\lambda_1}\mathbf{I} \right)^{-1}\Omega_r^\top \boldsymbol{\eta}_r) \\
       & = \left(\mathbf{J}\mathbf{U}^{-1}  + \frac{\lambda_2}{\lambda_1}\mathbf{I} \right)^{-1} \left(\frac{\lambda_2}{\lambda_1}\Phi_t^\top \boldsymbol{\epsilon}_t + \Omega_r^\top \boldsymbol{\eta}_r) \right) \\
   \end{align}
    Therefore, 
    \begin{align}
            T_{2a} &= \mathbf{U}^{-1}  T_{2a}^\prime \\ 
            &= \mathbf{U}^{-1}\left(\mathbf{J}\mathbf{U}^{-1}  + \frac{\lambda_2}{\lambda_1}\mathbf{I} \right)^{-1} \left(\frac{\lambda_2}{\lambda_1}\Phi_t^\top \boldsymbol{\epsilon}_t + \Omega_r^\top \boldsymbol{\eta}_r) \right) \\
            & = \left(\mathbf{J} + \frac{\lambda_2}{\lambda_1}\mathbf{U} \right)^{-1} \left(\frac{\lambda_2}{\lambda_1}\Phi_t^\top \boldsymbol{\epsilon}_t + \Omega_r^\top \boldsymbol{\eta}_r) \right) \\
            & = \left(\Omega_r^\top \Omega_r + \frac{\lambda_2}{\lambda_1}\Phi_t^\top\Phi_t + \lambda_2\mathbf{I}\right)^{-1}\left(\frac{\lambda_2}{\lambda_1}\Phi_t^\top \boldsymbol{\epsilon}_t + \Omega_r^\top \boldsymbol{\eta}_r) \right),
    \end{align}
    and 
    \begin{align} 
            T_2 &= \phi(\mathbf{x}) T_{2a} 
            \\
                & = \phi(\mathbf{x})\left(\Omega_r^\top \Omega_r + \frac{\lambda_2}{\lambda_1}\Phi_t^\top\Phi_t + \lambda_2\mathbf{I}\right)^{-1}\left(\frac{\lambda_2}{\lambda_1}\Phi_t^\top \boldsymbol{\epsilon}_t + \Omega_r^\top \boldsymbol{\eta}_r) \right)
                \\
                & \le \resizebox{0.9\textwidth}{!}{$ \norm{\phi(\mathbf{x}) \left(\Omega_r^\top \Omega_r + \frac{\lambda_2}{\lambda_1}\Phi_t^\top\Phi_t + \lambda_2\mathbf{I}\right)^{-1/2}}_{k_\textup{NTK-PINN}} \norm{\left(\Omega_r^\top \Omega_r + \frac{\lambda_2}{\lambda_1}\Phi_t^\top\Phi_t + \lambda_2\mathbf{I}\right)^{-1/2}\left(\frac{\lambda_2}{\lambda_1}\Phi_t^\top \boldsymbol{\epsilon}_t + \Omega_r^\top \boldsymbol{\eta}_r) \right)}_{k_\textup{NTK-PINN}}$}
                \\
                & = \resizebox{0.9\textwidth}{!}{$\sqrt{\phi(\mathbf{x})^\top\left(\Omega_r^\top \Omega_r + \frac{\lambda_2}{\lambda_1}\Phi_t^\top\Phi_t + \lambda_2\mathbf{I}\right)^{-1}\phi(\mathbf{x})} \norm{\left(\Omega_r^\top \Omega_r + \frac{\lambda_2}{\lambda_1}\Phi_t^\top\Phi_t + \lambda_2\mathbf{I}\right)^{-1/2}\left(\frac{\lambda_2}{\lambda_1}\Phi_t^\top \boldsymbol{\epsilon}_t + \Omega_r^\top \boldsymbol{\eta}_r) \right)}_{k_\textup{NTK-PINN}}$}
                \\
                &=\frac{\sigma_t^f(\mathbf{x})}{\lambda_2}\norm{\left(\Omega_r^\top \Omega_r + \frac{\lambda_2}{\lambda_1}\Phi_t^\top\Phi_t + \lambda_2\mathbf{I}\right)^{-1/2}\left(\frac{\lambda_2}{\lambda_1}\Phi_t^\top \boldsymbol{\epsilon}_t + \Omega_r^\top \boldsymbol{\eta}_r \right)}_{k_\textup{NTK-PINN}} 
                \\
                &=\frac{\sigma_t^f(\mathbf{x})}{\lambda_2}\norm{\left(\Omega_r^\top \Omega_r + \frac{\lambda_2}{\lambda_1}\Phi_t^\top\Phi_t + \lambda_2\mathbf{I}\right)^{-1/2}\left(\frac{\lambda_2}{\lambda_1}\Phi_t^\top \boldsymbol{\epsilon}_t \right) + \left(\Omega_r^\top \Omega_r + \frac{\lambda_2}{\lambda_1}\Phi_t^\top\Phi_t + \lambda_2\mathbf{I}\right)^{-1/2}\left( \Omega_r^\top \boldsymbol{\eta}_r \right)}_{k_\textup{NTK-PINN}} 
                \\
                & \le \resizebox{0.9\textwidth}{!}{$\frac{\sigma_t^f(\mathbf{x})}{\lambda_2} \left( \norm{\left(\Omega_r^\top \Omega_r + \frac{\lambda_2}{\lambda_1}\Phi_t^\top\Phi_t + \lambda_2\mathbf{I}\right)^{-1/2}\left(\frac{\lambda_2}{\lambda_1}\Phi_t^\top \boldsymbol{\epsilon}_t \right)}_{k_\textup{NTK-PINN}} + \norm{\left(\Omega_r^\top \Omega_r + \frac{\lambda_2}{\lambda_1}\Phi_t^\top\Phi_t + \lambda_2\mathbf{I}\right)^{-1/2}\left( \Omega_r^\top \boldsymbol{\eta}_r \right)}_{k_\textup{NTK-PINN}} \right)$}
                \\ 
                & = \resizebox{0.9\textwidth}{!}{$\frac{\sigma_t^f(\mathbf{x})}{\lambda_2}\left[ \frac{\lambda_2}{\lambda_1}\sqrt{\left(\Phi_t^\top \boldsymbol{\epsilon}_t \right)^\top\left(\Omega_r^\top \Omega_r + \frac{\lambda_2}{\lambda_1}\Phi_t^\top\Phi_t + \lambda_2\mathbf{I}\right)^{-1} 
                 \left(\Phi_t^\top \boldsymbol{\epsilon}_t\right)} + \sqrt{\left(\Omega_r^\top \boldsymbol{\eta}_r \right)^\top\left(\Omega_r^\top \Omega_r + \frac{\lambda_2}{\lambda_1}\Phi_t^\top\Phi_t + \lambda_2\mathbf{I}\right)^{-1} 
                 \left(\Omega_r^\top \boldsymbol{\eta}_r\right)} \right]$} 
                 \\
                 &= \frac{\sigma_t^f(\mathbf{x})}{\lambda_2} \left(\frac{\lambda_2}{\lambda_1} \norm{\Phi_t^\top\epsilon_t}_{\left(\Omega_r^\top \Omega_r + \frac{\lambda_2}{\lambda_1}\Phi_t^\top\Phi_t + \lambda_2\mathbf{I} \right)^{-1}} +\norm{\Omega_r^\top\epsilon_r}_{\left(\Omega_r^\top \Omega_r + \frac{\lambda_2}{\lambda_1}\Phi_t^\top\Phi_t + \lambda_2\mathbf{I} \right)^{-1}} \right) 
                 \\
                 & \le \frac{\sigma_t^f(\mathbf{x})}{\lambda_2} \Bigg(\frac{\lambda_2} {\lambda_1} \sqrt{2R_1^2 \log(\frac{\det(\Omega_r^\top \Omega_r + \frac{\lambda_2}{\lambda_1}\Phi_t^\top\Phi_t + \lambda_2\mathbf{I})^{1/2}\det(\Omega_r^\top \Omega_r + \lambda_2\mathbf{I})^{-1/2}}{\delta})}  
                 \\
                 & \;\;\;\;\;\;\;\;\;\;\;\;\;\;\;\;\;\; + \sqrt{2R_2^2 \log(\frac{\det(\Omega_r^\top \Omega_r + \frac{\lambda_2}{\lambda_1}\Phi_t^\top\Phi_t + \lambda_2\mathbf{I})^{1/2}\det(\frac{\lambda_2}{\lambda_1}\Phi_t^\top\Phi_t + \lambda_2\mathbf{I})^{-1/2}}{\delta})} \Bigg) \\
                 & \resizebox{0.95\textwidth}{!}{$\le \frac{\sigma_t^f(\mathbf{x})}{\lambda_2} \left(\sqrt{\left(\frac{\lambda_2}{\lambda_1}R_1\right)^2 + R_2^2} \sqrt{2\log(\frac{\det(\Omega_r^\top \Omega_r + \frac{\lambda_2}{\lambda_1}\Phi_t^\top\Phi_t + \lambda_2\mathbf{I})\det(\Omega_r^\top\Omega_r + \lambda_2\mathbf{I})^{-1/2}\det(\frac{\lambda_2}{\lambda_1}\Phi_t^\top\Phi_t + \lambda_2\mathbf{I})^{-1/2}}{\delta})}\right)$} 
                 \\
                 & = \frac{\sigma_t^f(\mathbf{x})}{\lambda_2} \left(\sqrt{\left(\frac{\lambda_2}{\lambda_1}R_1\right)^2 + R_2^2} \sqrt{2\log(\frac{\det(\Omega_r^\top \Omega_r + \frac{\lambda_2}{\lambda_1}\Phi_t^\top\Phi_t + \lambda_2\mathbf{I})}{\det(\Omega_r^\top\Omega_r + \lambda_2\mathbf{I})^{1/2}\det(\frac{\lambda_2}{\lambda_1}\Phi_t^\top\Phi_t + \lambda_2\mathbf{I})^{1/2}}) + \log(\frac{1}{\delta})}\right) 
                 \\
                 & = \frac{\sigma_t^f(\mathbf{x})}{\lambda_2} \left(\sqrt{\left(\frac{\lambda_2}{\lambda_1}R_1\right)^2 + R_2^2} \sqrt{2\log(\frac{\det( \frac{\Omega_r^\top \Omega_r}{\lambda_2} + \frac{\Phi_t^\top\Phi_t}{\lambda_1} + \mathbf{I})}{\det( \frac{\Omega_r^\top\Omega_r}{\lambda_2} + \mathbf{I})^{1/2}\det(\frac{\Phi_t^\top\Phi_t}{\lambda_1} + \mathbf{I})^{1/2}}) + \log(\frac{1}{\delta})}\right) 
                 \\
                 & =  \resizebox{0.9\textwidth}{!}{$\frac{\sigma_t^f(\mathbf{x})}{\lambda_2} \sqrt{\left(\frac{\lambda_2}{\lambda_1}R_1\right)^2 + R_2^2} \sqrt{2 \left[\log \det(\frac{\Phi_t^\top \Phi_t}{\lambda_1} + \frac{\Omega_r^\top \Omega_r}{\lambda_2} + \mathbf{I})^{1/2}  - \log (\frac{\det(\frac{\Phi_t^\top \Phi_t}{\lambda_1} + \mathbf{I})^{1/2}\det(\frac{\Omega_r^\top \Omega_r}{\lambda_2} + \mathbf{I})^{1/2}}{\det(\frac{\Phi_t^\top \Phi_t}{\lambda_1} + \frac{\Omega_r^\top \Omega_r}{\lambda_2} + \mathbf{I})^{1/2}}) \right] + \log(\frac{1}{\delta})} $}
                 \\
                 & \le \frac{\sigma_t^f(\mathbf{x})}{\lambda_2} \sqrt{\left(\frac{\lambda_2}{\lambda_1}R_1\right)^2 + R_2^2} \sqrt{\frac{1}{2}\log\det (\frac{\mathbf{K}_{uu}}{\lambda_1} + \mathbf{I}) - 2 I(f; \mathbf{Y}_t; \mathbf{U}_r) + \frac{N_r L^2}{2(1+ \rho_{min}(\mathbf{K}_{uu})/\lambda)} + \log(\frac{1}{\delta})} 
                 \\
                 & \le \frac{\sigma_t^f(\mathbf{x})}{\lambda_2} \sqrt{\left(\frac{\lambda_2}{\lambda_1}R_1\right)^2 + R_2^2} \sqrt{2\gamma_t - 2 I(f; \mathbf{Y}_t; \mathbf{U}_r) + \mathcal{O}(1) + \log(\frac{1}{\delta})} 
                 \\
                 & \le \sigma_t^f(\mathbf{x}) \sqrt{\left(\frac{R_1}{\lambda_1}\right)^2 + \left(\frac{R_2}{\lambda_2}\right)^2} \sqrt{2\gamma_t - 2 I(f; \mathbf{Y}_t; \mathbf{U}_r) + \mathcal{O}(1) + \log(\frac{1}{\delta})}
    \end{align}
The second inequality is based on triangle inequality while the third inequality relies on Lemma \ref{lemma:self-normalized_bound}. The fourth inequality employs the Cauchy-Schwarz inequality and the fifth inequality is based on Lemma \ref{lemma:IG_with_pde_upper_bound}.The final inequality stems from the choice of the number of PDE data points, denoted as $N_r$ as stated in Lemma \ref{lemma:confidence_bound}. We also utilizes the equation of $\left(\sigma_t^f\right)^2(\mathbf{x})$ at line \ref{Eqn:sigma_f_by_V}.
\end{proof}

\subsection{Proof of Theorem \ref{theorem:regret_bound}}
\RegretBound*

\paragraph{Proof sketch for Theorem \ref{theorem:regret_bound}} 
The proof strategy builds upon techniques from the regret bound of GP-TS introduced by \cite{chowdhury2017kernelized}. To analyze continuous actions in our context, we employ a discretization technique. At each time $t$, we utilize a discretization set $\mathcal{D}_t \subset \mathcal{D}$ such that $\lvert f(\mathbf{x}) - f([\mathbf{x}]_t) \rvert \leq \frac{1}{t^2}$ where $[\mathbf{x}]_t \in \mathcal{D}_t$ is the closest point to $\mathbf{x} \in \mathcal{D}$. We bound the regret of our proposed algorithm by starting with the instantaneous regret $r_t = f(\mathbf{x}^*) - f(\mathbf{x}_t)$ can be decomposed to $r_t = [f(\mathbf{x}^*) - f([\mathbf{x}^*]_t)] + [f([\mathbf{x}^*]_t) - f(\mathbf{x}_t)] \le \frac{1}{t^2} + \Delta_t(\mathbf{x}_t)$, where $[\mathbf{x}^*]_t \in \mathcal{D}_t$ is the closest point to optimal point $\mathbf{x}^*$. To bound the regret, it suffices to bound $\sum_{t=1}^t \Delta_t(\mathbf{x}_t)$. We partition the action space $\mathcal{D}_t$ into two sets: the \emph{saturated} set, denoted as $\mathcal{S}_t$, where $\Delta_t(\mathbf{x}) \ge c_t \sigma_t^f(\mathbf{x}), \mathbf{x} \in \mathcal{D}_t$, and the \emph{unsaturated} set, comprising points where $\Delta_t(\mathbf{x}) < c_t \sigma_t^f(\mathbf{x})$.  A crucial aspect of the proof involves demonstrating that the probability of selecting unsaturated points is sufficiently high. \emph{Here, our confidence bound in Lemma \ref{lemma:confidence_bound} comes into play to choose $c_T$. The cumulative regret bound $R_T = \sum_{i=1}^T r_t$ is then upper-bounded and we achieve the final expression as stated in Theorem \ref{theorem:regret_bound}}. 

\textbf{To prove Theorem \ref{theorem:regret_bound}, we start by proving the following lemmas:}
\begin{sublemma}
\label{lemma:var_sum_bound}
    Let $\mathbf{x}_1, \mathbf{x}_2,\dots, \mathbf{x}_T$ be the points selected by the Algorithms 1, then:
    \[\sum_{i=1}^T \sigma_t^f(\mathbf{x}) \le \sqrt{2T \left(\gamma_T - I_0 + \frac{N_r L^2}{2(1+ \rho_{min}(\mathbf{K}_{uu})/\lambda_1)}\right)}, \]
    where $\mathbf{K}_{uu}, \mathbf{K}_{ur}, \mathbf{K}_{ru}$ and $\mathbf{K}_{rr}$ are defined as in Lemma \ref{lemma:PINN_mean_cov} and $I_0 =\frac{1}{2}\log \frac{\det(\mathbf{K}_{rr} + \lambda_2\mathbf{I})}{\det(\mathbf{K}_{rr} + \lambda_2\mathbf{I} - \mathbf{K}_{ru} \mathbf{K}_{uu}^{-1} \mathbf{K}_{ur})}$
\end{sublemma}
\begin{proof}[Proof of Lemma \ref{lemma:var_sum_bound}]
conditioned on observations vector $\mathbf{Y}_{1:s-1}$  observed at points $\mathcal{A}_{s-1} \in \mathcal{D}$, and the PDE vector $\mathbf{U}_r = \mathbf{U}_{1:N_r}$, , the reward $y_s$ at round $s$ observed at
$\mathbf{x}_s$ is believed to follow the distribution $\mathcal{N}\left(\mu_s^f(\mathbf{x}_s); \nu_s^2[\lambda_1 + \left(\sigma_s^f\right)^2(\mathbf{x}_s)]\right)$, which gives:
\begin{align}
H(y_s \rvert \mathbf{Y}_{1:s-1}, \mathbf{U}_r) &= \frac{1}{2} \log\left[ 2\pi e \nu_s^2\left(\lambda_1 + \left(\sigma_s^f\right)^2(\mathbf{x}) \right) \right]\\
&=\frac{1}{2}\log(2\pi e \lambda_1\nu_s^2) + \frac{1}{2} \log \left[1+ \frac{\left(\sigma_s^f\right)^2(\mathbf{x})}{\lambda_1}\right].
\end{align}s
Now by chain rule of entropy function $H$: 
\begin{align}
    H(\mathbf{Y}_t; \mathbf{U}_r) &= H(y_t; y_{t-1};\dots, y_1; \mathbf{U}_r)
    \\
    &= H(\mathbf{U}_r) + H(y_1 \rvert \mathbf{U}_r) + H(y_2\rvert y_1, \mathbf{U}_r) + \dots + H (y_t \rvert \mathbf{y}_{1:t-1}, \mathbf{U}_r)
    \\
    & = H(\mathbf{U}_r) + \sum_{s=1}^t H(y_s \rvert \mathbf{y}_{1:s-1}, \mathbf{U}_r)
    \\ 
    \label{Eqn:joint_entropy_Yt_Ur}
    & = H(\mathbf{U}_r) + \frac{t}{2}\log(2\pi e \lambda_1\nu_s^2) + \frac{1}{2}\sum_{s=1}^t \log \left[1+ \frac{\left(\sigma_s^f\right)^2(\mathbf{x})}{\lambda_1}\right] \\
\end{align}
Next, by definition of interaction information of three random variables $f, \mathbf{Y}_t$ and $\mathbf{U}_r$, we have: 
    \begin{align}
        I(f; \mathbf{Y}_t; \mathbf{U}_r) &= H(f) + H(\mathbf{Y}_t) + H(\mathbf{U}_r) - [H(f; \mathbf{Y}_t) + H(\mathbf{Y}_t; \mathbf{U}_r) + H(f; \mathbf{U}_r)] + H (f; \mathbf{Y}_t; \mathbf{U}_r)
        \\
        &= H(f) - H(f \rvert \mathbf{Y}_t) - H(f \rvert \mathbf{U}_r) + H(f \rvert \mathbf{Y}_t; \mathbf{U}_r)
    \end{align}
    Therefore,  
    \begin{align}
       & H(f) - H(f \rvert \mathbf{Y}_t; \mathbf{U}_r) 
       \\
       & = I(f;\mathbf{Y}_t; \mathbf{U}_r) + H(f \rvert \mathbf{Y}_t)  + H(f \rvert \mathbf{U}_r) - 2 H(f \rvert \mathbf{Y}_t;\mathbf{U}_r)
       \\
        & = I(f;\mathbf{Y}_t; \mathbf{U}_r) + H(f \rvert \mathbf{Y}_t) - H(f \rvert \mathbf{Y}_t; \mathbf{U}_r)  + H(f \rvert \mathbf{U}_r) - H(f \rvert \mathbf{Y}_t; \mathbf{U}_r) 
        \\
        &= I(f;\mathbf{Y}_t; \mathbf{U}_r) + I(f;\mathbf{U}_r \rvert \mathbf{Y}_t)  +I(f;\mathbf{Y}_t \rvert \mathbf{U}_r)
        \\
        &= I(f;\mathbf{Y}_t; \mathbf{U}_r)  + \frac{1}{2} \log \frac{\det(\Phi_t \left(\frac{\Omega_r^\top \Omega_r}{\lambda_2} + \mathbf{I}\right)^{-1} \Phi_t^\top)}{\det(\Phi_t \left(\frac{\Omega_r^\top\Omega_r}{\lambda_2}+\frac{\Phi_t^\top \Phi_t} {\lambda_1} + \mathbf{I}\right)^{-1} \Phi_t^\top)} + \frac{1}{2} \log \frac{\det(\Phi_t \left(\frac{\Phi_t^\top \Phi_t}{\lambda_1} + \mathbf{I}\right)^{-1} \Phi_t^\top)}{\det(\Phi_t \left(\frac{\Omega_r^\top\Omega_r}{\lambda_2}+\frac{\Phi_t^\top \Phi_t} {\lambda_1} + \mathbf{I}\right)^{-1} \Phi_t^\top)}
        \\
        &= I(f;\mathbf{Y}_t; \mathbf{U}_r)  + \frac{1}{2} \log \frac{\det \left(\frac{\Omega_r^\top \Omega_r}{\lambda_2} + \mathbf{I}\right)^{-1}}{\det \left(\frac{\Omega_r^\top\Omega_r}{\lambda_2}+\frac{\Phi_t^\top \Phi_t} {\lambda_1} + \mathbf{I}\right)^{-1}} + \frac{1}{2} \log \frac{\det\left(\frac{\Phi_t^\top \Phi_t}{\lambda_1} + \mathbf{I}\right)^{-1}}{\det \left(\frac{\Omega_r^\top\Omega_r}{\lambda_2}+\frac{\Phi_t^\top \Phi_t} {\lambda_1} + \mathbf{I}\right)^{-1}}
        \\
        &= \frac{1}{2} \left[ \log \frac{\det(\frac{\Phi_t^\top \Phi_t}{\lambda_1} + \mathbf{I})\det(\frac{\Omega_r^\top \Omega_r}{\lambda_2} + \mathbf{I})}{\det(\frac{\Phi_t^\top \Phi_t}{\lambda_1} + \frac{\Omega_r^\top \Omega_r}{\lambda_2} + \mathbf{I})} + \log \frac{\det \left(\frac{\Omega_r^\top \Omega_r}{\lambda_2} + \mathbf{I}\right)^{-1}}{\det \left(\frac{\Omega_r^\top\Omega_r}{\lambda_2}+\frac{\Phi_t^\top \Phi_t} {\lambda_1} + \mathbf{I}\right)^{-1}} + \log \frac{\det\left(\frac{\Phi_t^\top \Phi_t}{\lambda_1} + \mathbf{I}\right)^{-1}}{\det \left(\frac{\Omega_r^\top\Omega_r}{\lambda_2}+\frac{\Phi_t^\top \Phi_t} {\lambda_1} + \mathbf{I}\right)^{-1}} \right]
        \\
        &=\frac{1}{2}\log \det(\frac{\Phi_t^\top \Phi_t}{\lambda_1} + \frac{\Omega_r^\top \Omega_r}{\lambda_2} + \mathbf{I})
        \\
\label{Eqn:H(f)_minus_H(f_give_Yt_Ur)}
        &=
        H(\mathbf{Y}_t; \mathbf{U}_r) - H(\mathbf{Y}_t; \mathbf{U}_r \rvert f) 
    \end{align}
    The fourth equation uses the expression of $(f;\mathbf{Y}_t\rvert\mathbf{U}_r)$ in Eqn. \ref{Eqn:conditional_information_f_and_Yt_give_Ur} and the fifth equation directly uses the result in Corollary \ref{corollary:det_divison_corollary}.
    The sixth equation is obtained by using the expression of $I(f;\mathbf{Y}_t; \mathbf{U}_r)$  in Lemma \ref{lemma:interaction_information_formula} while the last equation uses the symmetry of mutual information.
    Using similar technique as showed in the proof of Lemma \ref{lemma:interaction_information_formula}, we have $\textup{Cov}(\mathbf{Y}_t; \mathbf{U}_r \rvert f) = \mathbf{K}_{rr} + \lambda_2 \mathbf{I} -  \mathbf{K}_{ru} \mathbf{K}_{uu}^{-1} \mathbf{K}_{ur}$. Combining Eqn \ref{Eqn:joint_entropy_Yt_Ur} and Eqn \ref{Eqn:H(f)_minus_H(f_give_Yt_Ur)}, we have 
    \begin{align}
        H(\mathbf{U}_r) + \frac{t}{2}\log(2\pi e \lambda_1\nu_s^2) + \frac{1}{2}\sum_{s=1}^t \log \left[1+ \frac{\left(\sigma_s^f\right)^2(\mathbf{x})}{\lambda_1}\right] - H(\mathbf{Y}_t; \mathbf{U}_r \rvert f) =\frac{1}{2}\log \det(\frac{\Phi_t^\top \Phi_t}{\lambda_1} + \frac{\Omega_r^\top \Omega_r}{\lambda_2} + \mathbf{I})
    \end{align}
    or, 
    \begin{align}
      \frac{1}{2}\sum_{s=1}^t \log \left[1+ \frac{\left(\sigma_s^f\right)^2(\mathbf{x})}{\lambda_1}\right] &=  \frac{1}{2}\log \det(\frac{\Phi_t^\top \Phi_t}{\lambda_1} + \frac{\Omega_r^\top \Omega_r}{\lambda_2} + \mathbf{I}) - \left[H(\mathbf{U}_r) - H(\mathbf{Y}_t; \mathbf{U}_r \rvert f) \right] - \frac{t}{2}\log(2\pi e \lambda_1\nu_s^2) 
      \\
      &= \frac{1}{2}\log \det(\frac{\Phi_t^\top \Phi_t}{\lambda_1} + \frac{\Omega_r^\top \Omega_r}{\lambda_2} + \mathbf{I}) - \frac{1}{2} \log \frac{\det (\mathbf{K}_{rr} + \lambda_2\mathbf{I})}{\det(\mathbf{K}_{rr} + \lambda_2 \mathbf{I} -  \mathbf{K}_{ru} \mathbf{K}_{uu}^{-1} \mathbf{K}_{ur})}
    \end{align}

    Finally, we have
    \begin{align}
        \sum_{i=1}^T \sigma_t^f(\mathbf{x}) & \le \sqrt{T} \sqrt{ \sum_{i=1}^T \left(\sigma_t^f(\mathbf{x})\right)^2}
        \\
        & \le \sqrt{T} \sqrt{\sum_{i=1}^T 2\lambda_1\log \left[\frac{\left(\sigma_s^f\right)^2(\mathbf{x})}{\lambda_1} +1 \right]} 
        \\
        & \le \sqrt{2T\lambda_1} \sqrt{\log \det(\frac{\Phi_t^\top \Phi_t}{\lambda_1} + \frac{\Omega_r^\top \Omega_r}{\lambda_2} + \mathbf{I}) - \log \frac{\det (\mathbf{K}_{rr} + \lambda_2\mathbf{I})}{\det(\mathbf{K}_{rr} + \lambda_2 \mathbf{I} -  \mathbf{K}_{ru} \mathbf{K}_{uu}^{-1} \mathbf{K}_{ur})}}
        \\
        & \le \sqrt{2T\left(\gamma_T - I_0 + \frac{N_r L^2}{2(1+ \rho_{min}(\mathbf{K}_{uu})/\lambda_1)}\right)},
    \end{align}
    where $I_0= \frac{\det (\mathbf{K}_{rr} + \lambda_2\mathbf{I})}{\det(\mathbf{K}_{rr} + \lambda_2 \mathbf{I} -  \mathbf{K}_{ru} \mathbf{K}_{uu}^{-1} \mathbf{K}_{ur})}$. 
    The first inequality uses the Cauchy-Swcharz inequality, while the second inequality uses the fact that $x \le 2\log(x+1), \forall \; 0 \le x \le 1$. The third inequality uses the result from Lemma \ref{lemma:IG_with_pde_upper_bound}. Now the result follows by choosing $\lambda_1 = 1 + 1/T$.
\end{proof}
\begin{sublemma}[Lemma 5 \cite{chowdhury2017kernelized}]
\label{lemma:sampling_bound}
For any $t \in [T]$, and any finite subset $\mathcal{D}_t \subset\ \mathcal{D}$,  pick $\widetilde{c}_t =  \sqrt{4\log t + 2 \log \ \lvert \mathcal D_t \rvert}$. Then  
\[\lvert \widetilde{f}_t(\mathbf{x}) - \mu_{t-1}^f(\mathbf{x}) \rvert \leq \nu_t \widetilde{c}_t \sigma_t^f(\mathbf{x}), \forall \mathbf{x} \in \mathcal{D}_t, \] 
holds with probability $\geq  1-t^{-2}$, and $\widetilde{f}_t(\mathbf{x}) = \nu_t h(\mathbf{x}; \boldsymbol{\theta}_{t-1})$ is the acquisition value stated in Algorithm 1.   
\end{sublemma}
\begin{sublemma} [Lemma 13, \cite{chowdhury2017kernelized}]
\label{lemma:regret}
Given any $\delta \in (0, 1)$, with probability at least $1 - \delta$, 
\begin{align}
    R_T = 
    \frac{11c_T}{p}\sum_{t=1}^T \sigma_t^f(\mathbf{x}_t) + \frac{(2B+1)\pi^2}{6} + \frac{(4B+11)c_T}{p} \sqrt{2T\log(2/\delta)} ,
\end{align}
where $p=\frac{1}{4e\pi}$ and $c_T$ is a time-dependent factor. 
\end{sublemma}
Now we are ready to prove the Theorem \ref{theorem:regret_bound}.
To perform analysis for continuous search space, we use a discretization technique. At each time $t$, we use a discretization $\mathcal{D}_t^{dis} \subset \mathcal{D} \subset [0, r]^d$, where $d$ is the dimension, which satisfies the property that $\lvert f(\mathbf{x}) - f([\mathbf{x}]_t) \rvert \leq \frac{1}{t^2}$ where $[\mathbf{x}]_t \in \mathcal{D}_t^{dis}$ is the closest point to $\mathbf{x} \in \mathcal{D}$. Then we choose $\mathcal{D}_t^{dis}$ 
with size $\lvert \mathcal{D}_t^{dis} \rvert = \left( BC_{\text{lip}}rdt^2 \right)^d$ that satisfies $\norm{\mathbf{x} - [\mathbf{x}]_t}_1 \leq \frac{rd}{BC_{\text{lip}}rdt^2} = \frac{1}{BC_{\text{lip}}t^2}$ for all $\mathbf{x} \in \mathcal{D}$,
where $C_{\text{lip}}= \underset{\mathbf{x} \in \mathcal{D}} {\supremum} \underset{j \in [d]} {\supremum} \left( \frac{\partial^2 k_\text{NTK-PINN}(\mathbf{p},\mathbf{q})}{\partial \mathbf{p}_j \mathbf{q}_j} \lvert \mathbf{p}=\mathbf{q}=\mathbf{x} \right)$. This implies, for all $\mathbf{x} \in \mathcal{D}$:
\begin{equation}
\label{eqn:rkhs_lipschitz}
    \lvert f(\mathbf{x}) - f([\mathbf{x}]_t) \rvert \leq \norm{f}_{\mathcal{H}_{k_\textup{NTK-PINN}}} C_{\text{lip}} \norm{\mathbf{x} - [\mathbf{x}]_t}_1 \leq BC_{\text{lip}} \frac{1}{BC_{\text{lip}}t^2} =  1/t^2, 
\end{equation}
is Lipschitz continuous of any $f \in \mathcal{H}_{k_\textup{NTK}}(\mathcal{D})$ with Lipschitz constant $B C_{\text{lip}}$, where we have used the inequality $\norm{f}_{\mathcal{H}_{k_\textup{NTK-PINN}}} \leq B $  which is our assumption about function $f$.

Then, we define two events as follows:
Define $E^f(t)$ as the event that for all $\mathbf{x} \in \mathcal{D}$
\begin{align} 
\lvert \mu_{t-1}^f(\mathbf{x}) - f(\mathbf{x}) \rvert \le \nu_t \sigma_{t-1}^f (\mathbf{x}), 
\end{align}
and $E^{f_t}(t)$ as the event that for all $\mathbf{x} \in \mathcal{D}$
\begin{align}
    \lvert \widetilde{f}_t(\mathbf{x}) - \mu_{t-1}^f(\mathbf{x})\rvert \le \nu_t \widetilde{c}_t\sigma_{t-1}^f(x)
\end{align}

Lemma \ref{lemma:sampling_bound} implies that event $E^{f_t}(t)$ holds w.p $1-1/t^2$, while our Lemma \ref{lemma:confidence_bound} implies that event $E^f(t)$ holds w.p $1-\delta$ with $\delta \in (0,1)$. Further, Lemma \ref{lemma:confidence_bound} also hint to choose $\nu_t = \widetilde{R} \sqrt{2\gamma_t - 2 I(f; \mathbf{Y}_t; \mathbf{U}_r) + \mathcal{O}(1) + \log(\frac{1}{\delta})}$, where $\widetilde{R} = \sqrt{\left(\frac{R_1}{\lambda_1}\right)^2 + \left(\frac{R_2}{\lambda_2}\right)^2}$. 

Then, let $c_t = \nu_t \widetilde{c}_t + \nu_t = \nu_t (1+\widetilde{c}_t)$. Following the Lemma \ref{lemma:regret}, with probability at least $1-\delta$, the cummulative regret can be bounded as: 
\begin{align}
    R_T &= \mathcal{O}\left(c_T\left(\sum_{t=1}^T \sigma_t^f(\mathbf{x}) + \sqrt{T\log(2/\delta)}\right)\right)
    \\
    &= \mathcal{O}\left( \left[B + \widetilde{R}\sqrt{2\gamma_T - 2I(f;\mathbf{Y}_T; \mathbf{U}_r)+\log(1/\delta)} \right] \left[1 +  \sqrt{4 \log T + 2d\log BdT} \right] \left[ \sqrt{T(\gamma_T - I_0)} + \sqrt{T\log(2/\delta)} \right] \right)
    \\
    &= \mathcal{O}\left( \left[B + \widetilde{R}\sqrt{2\gamma_T - 2I(f;\mathbf{Y}_T; \mathbf{U}_r)+\log(1/\delta)} \right] \left[1 +  \sqrt{d\log BdT}\right] \left[\sqrt{T(\gamma_T - I_0)} + \sqrt{T\log(2/\delta)}\right] \right) 
    \\
    &= \mathcal{O}\left( \sqrt{T} \left[B + \widetilde{R}\sqrt{2\gamma_T - 2I(f;\mathbf{Y}_T; \mathbf{U}_r)+\log(1/\delta)} \right]  \left[1 +  \sqrt{d\log BdT}\right]  \sqrt{\gamma_T - I_0 + \log(2/\delta)}   \right)
    \\
    &= \mathcal{O}\left( \sqrt{T} \left(1 +  \sqrt{d\log BdT}\right) \left(B + \widetilde{R}\sqrt{2\gamma_T - 2I(f;\mathbf{Y}_T; \mathbf{U}_r)+\log(1/\delta)} \right) \sqrt{\gamma_T - I_0 + \log(2/\delta)}   \right) 
    \\
    &= \mathcal{O}\left( \sqrt{T d\log BdT} \left(B \sqrt{\gamma_T - I_0 + \log(2/\delta)} + \widetilde{R}\sqrt{\gamma_T^2 - \gamma_T \left[I(f;\mathbf{Y}_T; \mathbf{U}_r) + I_0 -\log(2/\delta)\right]} \right) \right) 
    \\
    &= \mathcal{O}\left( \sqrt{T d\log BdT} \left(B \sqrt{\gamma_T - I_0 + \log(2/\delta)} + \widetilde{R}\sqrt{\gamma_T}\sqrt{\gamma_T - \left[I(f;\mathbf{Y}_T; \mathbf{U}_r) + I_0 -\log(2/\delta)\right]} \right) \right)
    \\
    &= \mathcal{O}\left( \sqrt{T d\log BdT} \left[B \sqrt{\gamma_T - I_0 + \log(2/\delta)} + \widetilde{R}\sqrt{\gamma_T}\sqrt{\gamma_T - I(f;\mathbf{Y}_T; \mathbf{U}_r) - I_0 +\log(2/\delta)} \right] \right)
\end{align}
\pagebreak

\end{document}